\documentclass{clv2025}

\usepackage[utf8]{inputenc}
\usepackage[T1]{fontenc}
\jvol{04}
\jnum{07}
\jyear{2025}

\dochead{Position Paper} 

\usepackage{amsmath}
\usepackage{booktabs}
\usepackage{bookmark}
\hypersetup{hidelinks}
\usepackage{graphicx}
\usepackage{multirow}
\usepackage{tikz}
\usepackage{pinyin}
\usetikzlibrary{positioning, shapes, arrows.meta, fit, backgrounds}

\runningtitle{Towards Computational Chinese Paleography}
\runningauthor{Yiran Rex Ma}

\begin{document}

\title{Towards Computational Chinese Paleography}


\author{Yiran Rex Ma\thanks{This work is in progress, with the Peking University-ByteDance Digital Humanities Open Lab. Note that Chinese references are presented here in English translation only; for complete bibliographic details, please refer to the \LaTeX{} source.}}


\affilblock{
\affil{School of Humanities, Beijing Univeristy of Posts and Telecommunications\\\quad \email{mayiran@bupt.edu.cn}}
}

\maketitle

\begin{abstract}
Chinese paleography, the study of ancient Chinese writing, is undergoing a computational turn powered by artificial intelligence. This position paper charts the trajectory of this emerging field, arguing that it is evolving from automating isolated visual tasks to creating integrated digital ecosystems for scholarly research. We first map the landscape of digital resources, analyzing critical datasets for oracle bone, bronze, and bamboo slip scripts. The core of our analysis follows the field's methodological pipeline: from foundational visual processing (image restoration, character recognition), through contextual analysis (artifact rejoining, dating), to the advanced reasoning required for automated decipherment and human-AI collaboration. We examine the technological shift from classical computer vision to modern deep learning paradigms, including transformers and large multimodal models. Finally, we synthesize the field's core challenges—notably data scarcity and a disconnect between current AI capabilities and the holistic nature of humanistic inquiry—and advocate for a future research agenda focused on creating multimodal, few-shot, and human-centric systems to augment scholarly expertise.
\end{abstract}

\section{Introduction}
\label{sec:introduction}

Chinese paleography, the study of ancient Chinese writing systems, stands as a foundational discipline for investigating the origins and evolution of Chinese civilization, language, and history. As the world's only logographic script that has been continuously used from antiquity to the present day, it provides a unique and unbroken record of cultural and linguistic development. Having emerged over the past century from the traditions of classical epigraphy, it has matured into a vital interdisciplinary field, deeply interwoven with history, archaeology, linguistics, and philology. 

Its development has been continually propelled by a series of momentous archaeological discoveries, from the oracle bones of the Shang Dynasty to the bronze inscriptions of the Zhou and the bamboo and silk manuscripts of the War States, Qin, and Han periods\footnote{The dynasties mentioned are the Shang (c. 1600--1046 BCE), Zhou (c. 1046--256 BCE), Warring States (475--221 BCE), Qin (221--206 BCE), and Han (206 BCE--220 CE).} \cite{fudan_excavated_2024}. In recent decades, the field has entered a golden age, boosted by three key factors: an explosion in the quantity of unearthed materials, the increasingly systematic and high-quality collation of these resources, and a significant rise in the overall level of scholarly research \cite{huang_construction_2025}. 

This era has coincided with another transformative development: the maturation of artificial intelligence (AI) as a powerful tool for humanistic inquiry. The ambition to apply computational methods is not new, with experiments on oracle bones dating back to the 1970s and attempt to design pictographic code\cite{zh-li_computer_1996} and a steady stream of research since \cite{zh-wang_computer_2010-2,zh-wang_intelligent_2010,zh-wang_computer_2011}. However, it is the recent surge in AI capabilities that has truly begun to unlock this potential. The success of systems like Ithaca, which restores and attributes ancient Greek inscriptions with remarkable accuracy \cite{assael_restoring_2022}, and its successor Aeneas, which further enhances performance by integrating visual and textual data \cite{assael_contextualizing_2025}, has showcased the potential for AI to revolutionize the study of ancient texts.

This confluence of factors has created a set of challenges that together necessitate a computational turn. First, the sheer volume and fragmentary nature of unearthed materials demand new methods for comprehensive, high-level collation and organization, forming the bedrock of future research. Second, and more critically, the discipline faces fundamental research bottlenecks: the decipherment of difficult characters remains a formidable task, and there is a pressing need to move beyond fragmented analyses of new materials toward integrated, systematic studies that can reveal deeper historical and linguistic patterns \cite{huang_construction_2025}. While applying AI to the unique logographic system of ancient Chinese presents distinct challenges compared to alphabetic scripts \cite{zh-mo_application__2023}, the response to these pressures has been the emergence of \textbf{computational Chinese paleography}\cite{mo2022computational,huang_construction_2025}. This field has rapidly evolved from a narrow focus on task automation to a broader vision of building integrated digital research ecosystems, and this survey argues that its success will hinge not only on technical innovation but on a deep and respectful integration of computational methods with the rich traditions of humanistic inquiry\footnote{The advance of digital technology has reshaped paleography research, demanding greater digital literacy from scholars and leading to the development of numerous databases and online platforms \cite{zh-mo_daily_2025}. Although these digital infrastructures facilitate material retrieval, collection, and investigation, involving computational efforts, this survey will focus on (intelligent) computing applications in paleography, rather than the infrastructure itself.}.

This rapid development has led to a proliferation of new datasets, models, and applications. While several reviews have touched upon aspects of this field, they have often focused narrowly on specific tasks like image-level character recognition \cite{diao_ancient_2025} or a single script type, majorly oracle bone inscriptions (OBIs) \cite{li_comprehensive_2024, li_mitigating_2025}. Consequently, they have not fully captured the expanding, multimodal, AI-human-interactive, and interdisciplinary scope of the field. In contrast, this paper adopts a comprehensive, problem-oriented framework to analyze the full pipeline of computational paleography, aiming to articulate its current state and future potential. The key contributions of this work are as follows:
\begin{enumerate}
    \item \textbf{A Problem-Oriented Analysis of the Methodological Pipeline:} We present a comprehensive analysis structured around a problem-oriented pipeline, progressing from foundational visual analysis (e.g., image restoration, character recognition) to contextual tasks (e.g., artifact rejoining, dating) and culminating in advanced reasoning (e.g., automated decipherment). This framework consolidates the field's data infrastructure, highlighting the crucial shift from isolated character datasets to full-document, multimodal resources, while also mapping the primary computational challenges researchers currently face.
    \item \textbf{Analysis of the Field's Technological Trajectory:} We trace the evolution from classical feature engineering to modern deep learning, analyzing the application of key architectures to specific paleographical problems. This includes the use of CNNs for visual feature extraction, transformers for modeling bamboo slip texts, and diffusion models for generative decipherment. We particularly highlight the emerging trend of methods grounded in the logographic nature of Chinese script, such as radical-based recognition and compositional reasoning, which move beyond generic computer vision paradigms.
    \item \textbf{Synthesis of Core Challenges and a Forward-Looking Agenda:} We identify critical obstacles, including data scarcity and a fundamental disconnect between current AI, which focuses on graphic form, and holistic paleographical inquiry, which integrates phonological and contextual evidence. We then chart a research agenda that pivots from the goal of full automation to one of human-AI synergy. This future direction emphasizes developing multimodal, few-shot systems designed not to replace, but to augment scholarly expertise, acting as intelligent assistants for knowledge discovery and intellectual partners for overcoming research bottlenecks.
\end{enumerate}

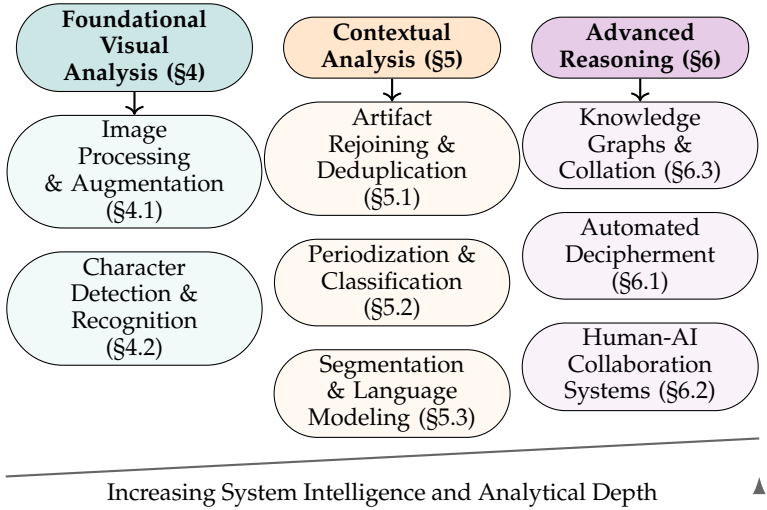
\begin{figure*}[t]
\centering
\begin{tikzpicture}[
    node distance=0.3cm and 0.4cm,
    base_stage/.style={draw, rounded rectangle, text centered, font=\small\bfseries, text width=2.4cm, minimum height=0.8cm, inner sep=2pt},
    base_task/.style={draw, rounded rectangle, text centered, font=\small, text width=2.4cm, minimum height=0.8cm, inner sep=2pt},
    stage1/.style={base_stage, fill=teal!20},
    task1/.style={base_task, fill=teal!5},
    stage2/.style={base_stage, fill=orange!20},
    task2/.style={base_task, fill=orange!5},
    stage3/.style={base_stage, fill=violet!20},
    task3/.style={base_task, fill=violet!5},
    axis/.style={-Latex, thick, draw=black!60}
]

\node (found) [stage1] {Foundational Visual Analysis (\S\ref{sec:visual_analysis})};
\node (imgproc) [task1, below=of found] {Image Processing \& Augmentation (\S\ref{sec:process})};
\node (recog) [task1, below=of imgproc] {Character Detection \& Recognition (\S\ref{sec:reg})};

\node (context) [stage2, right=of found] {Contextual Analysis (\S\ref{sec:contextual_analysis})};
\node (rejoin) [task2, below=of context] {Artifact Rejoining \& Deduplication (\S\ref{sec:rejoining})};
\node (dating) [task2, below=of rejoin] {Periodization \& Classification (\S\ref{sec:dating})};
\node (nlp) [task2, below=of dating] {Segmentation \& Language Modeling (\S\ref{sec:nlp})};

\node (advanced) [stage3, right=of context] {Advanced Reasoning (\S\ref{sec:advanced_reasoning})};
\node (kg) [task3, below=of advanced] {Knowledge Graphs \& Collation (\S\ref{sec:kg})};
\node (decipher) [task3, below=of kg] {Automated Decipherment (\S\ref{sec:decipherment})};
\node (collab) [task3, below=of decipher] {Human-AI Collaboration Systems (\S\ref{sec:collaboration})};

\draw[->, thick] (found) -- (imgproc);
\draw[->, thick] (context) -- (rejoin);
\draw[->, thick] (advanced) -- (kg);

\begin{scope}[on background layer]
    \node[fit=(found)(recog)(context)(nlp)(advanced)(collab), inner sep=0pt] (allnodes) {};
\end{scope}
\draw[axis] (allnodes.south west) ++(0,-0.5cm) -- (allnodes.south east) ++(0,-0.5cm) node[midway, below=2mm, font=\small] {Increasing System Intelligence and Analytical Depth};

\end{tikzpicture}
\caption{The methodological progression charted in this paper, from foundational visual processing to advanced reasoning systems. The full structure, including introductory and concluding sections, is detailed in the main text.}
\label{fig:structure}
\end{figure*}

Accordingly, this paper is structured to reflect a progression from foundational tasks to increasingly complex applications that demonstrate greater system intelligence (see Figure \ref{fig:structure}). We begin in \S\ref{sec:resources} by charting the landscape of digital resources. We then proceed through three main stages of computational work. In \S\ref{sec:visual_analysis}, we cover foundational visual analysis, from image processing (\S\ref{sec:process}) to character recognition (\S\ref{sec:reg}). In \S\ref{sec:contextual_analysis}, we examine the analysis of artifacts and texts in context, including fragment rejoining (\S\ref{sec:rejoining}), dating (\S\ref{sec:dating}), and language modeling (\S\ref{sec:nlp}). In \S\ref{sec:advanced_reasoning}, we explore the frontier of advanced reasoning and knowledge systems, covering knowledge graphs (\S\ref{sec:kg}), automated decipherment (\S\ref{sec:decipherment}), and human-AI collaboration (\S\ref{sec:collaboration}). Finally, we discuss the field's prevailing challenges in \S\ref{sec:discussion} and conclude in \S\ref{sec:conclusion}.

\section{From Classical Epigraphy to Computational Paleography}
\label{sec:background}

\subsection{The Scope, Materials, and Tasks of Chinese Paleography}
\label{sec:classical_paleography}

The historical trajectory of the Chinese script unfolds across several major stages. It begins with the Oracle Bone Inscriptions (\textit{ji\v{a} g\v{u} w\'{e}n}) of the late Shang Dynasty (c. 1600--1046 BCE), followed by the Bronze Inscriptions (\textit{j\={i}n w\'{e}n}) prominent during the Shang and Zhou Dynasties (c. 1600--256 BCE). The script of the Warring States period (475--221 BCE), often found on bamboo/wooden slips and silk (\textit{ji\v{a}n b\'{o}}). This was succeeded by the standardized Small Seal script (\textit{xi\v{a}o zhu\`{a}n}) of the Qin Dynasty (221--206 BCE), which in turn evolved into the Clerical script (\textit{l\`{i} sh\={u}}) of the Han Dynasty (206 BCE--220 CE). The process culminated in the Regular script (\textit{k\v{a}i sh\={u}}), which emerged around the end of the Han Dynasty and has remained the standard form (traditional Chinese) ever since (after with simplified Chinese).

While the script continued to evolve, the scope of Chinese paleography traditionally focuses on the study of ancient scripts from the Oracle Bone Inscriptions to ``the clerical scripts of the Han Dynasty that still retain characteristics of seal script''\cite{fudan_excavated_2024}. These ancient scripts are the medium for excavated documents. In its broadest sense, classical Chinese paleography encompasses both the study of the scripts themselves and the physical artifacts that bear them \cite{qiu2000chinese}. The terms ``ancient scripts'' and ``excavated documents'' are often used interchangeably in Chinese paleographic scholarship.\footnote{Chinese textual materials are broadly divided into excavated documents (\textit{ch\={u} t\v{u} wén xiàn}) and received/transmitted texts (\textit{chuán shì wén xiàn}), which are used with mutual corroboration in modern historiography\cite{chen_yinque_collected_works_2001}. While computation of received texts, such as segmentation\cite{hu_new_2021,zh-tang_ancient-chinese-word_2023,tang_incorporating_2023}, information extraction\cite{tang_chisiec_2024}, classical-modern translation, and culture analysis\cite{wang_evol_2024}, is undoubtedly valuable for paleographic studies, this survey focuses specifically on the challenges related to excavated documents and (majorly) the ancient scripts they contain.}

These materials can be classified based on several criteria. \textbf{Chronologically}, the scripts are often divided into four major periods: the Shang Dynasty, the Western Zhou Dynasty (1046--771 BCE), the Spring and Autumn (771--476 BCE) and Warring States (475--221 BCE) periods, and the Qin (221--206 BCE) and Han Dynasties. \textbf{By writing medium}, they are categorized by the materials on which the scripts are inscribed, including oracle bones, bronze vessels, bamboo and wooden slips, and silk manuscripts. For longer texts, they can also be classified \textbf{by content and genre}\footnote{Following the bibliographic system of the \textit{Hanshu Yiwenzhi} (``Treatise on Arts and Literature'' from \textit{The Book of Han}). Six Arts include: Rites, Music, Archery, Charioteering, Calligraphy, and Mathematics}, such as the Six Arts (\textit{li\`u y\`i}), philosophical schools (e.g. Confucianism and Daoism), poetry and rhapsodies, military strategy, numerology, and occult and practical arts.

The content of these documents evolved significantly over time. \textbf{Oracle Bone Inscriptions} from the Shang Dynasty primarily consist of divination records (\textit{b\v{u} c\'i}), with a smaller number of inscriptions for general record-keeping (\textit{j\`i sh\`i k\`e c\'i}). \textbf{Western Zhou bronze inscriptions} expand in scope to include records of merits and rewards, sales contracts, major state events, and legal cases. From the \textbf{Spring and Autumn period onward}, the content diversifies further to include sacrificial prayers, curses, and artisan's marks inscribed on objects. It was during this time, with the proliferation of bamboo and silk manuscripts in the late Warring States, Qin, and Han periods, that ``books'' in the modern sense began to appear in large numbers, containing the works of individual thinkers and philosophical schools. In \textbf{Warring States bamboo slips}, for instance, the state system produced a large volume of legal and administrative texts, while the private sphere generated numerous popular almanacs used for divination in daily life. 
 
As an interdisciplinary field rooted in epigraphy, Chinese paleography intersects deeply with history, archaeology, and linguistics. Its scholarly process can be understood as a hierarchy of tasks. The foundational work is \textbf{collation}, which involves the scientific documentation, transcription, and publication of unearthed materials. This further includes tasks like rejoining fragmented artifacts, identifying duplicate texts, authenticating documents to distinguish them from forgeries, and dating the materials.

The core intellectual challenge, however, lies in \textbf{decipherment and interpretation}. This is not merely a task of visual recognition but a holistic analytical process. Scholars must synthesize evidence from a character's graphic form, its presumed phonological properties, and its semantic role in context. One begins by seeking a plausible graphic form, tests it against the textual meaning, and refines the interpretation by appealing to grammar and phonology until a satisfactory reading is achieved \cite{mo2022computational}. The ultimate goal of this work is to enable deeper \textbf{thematic and synthetic research}, using the deciphered texts to reconstruct ancient history and society, and to systematically uncover the developmental laws of the Chinese script itself.

\subsection{The Unique Challenge of Chinese Paleography}
Typologically, the world's writing systems are broadly distinguished into phonographic systems and logographic systems\cite{liu1995script,gnanadesikan2017towards}. Ancient Chinese script, a unique and independently formed system, represents a particularly complex branch of the logographic family. It is more precisely defined as a \textbf{morphemo-syllabic script} \cite{qiu2000chinese}, where each character generally corresponds to a single morpheme and a single syllable. This contrasts sharply with phonographic scripts (e.g., Latin, Greek), whose basic units are a small, closed set of letters with no inherent meaning. The Chinese script, instead, employs tens of thousands of complex 
characters composed of symbols that can be semantic, phonetic, or arbitrary signs. Alternatively, it can also be seen as being built from hundreds of components (radicals), which combine in non-linear arrangements. 

While the language model is just as crucial for logographic as for phonographic scripts, the sheer scale of the visual recognition task makes the image modality an indispensable and formidable challenge \cite{zh-mo_multimodal_2024}. The composition of Chinese characters is explained by structural theories, with modern grammatology (i.e. ``scientific paleography'') favoring a framework of three character types: semantographs, loangraphs, and phonograms.\cite{qiu2000chinese}\footnote{The traditional ``Six Principles'' (\textit{liù sh\={u}}) theory developed by Xu Shen\cite{xu_shuowen_1963} includes pictographs (\textit{xiàng xíng}), deictic graphs (\textit{zh\v{i} shì}), syssemantographs (\textit{huì yì}), phonograms (\textit{xíng sh\={e}ng}), derivative cognates (\textit{zhu\v{a}n zhù}), and loangraphs (\textit{ji\v{a} jiè}). The modern ``Three Principles'' (\textit{s\={a}n sh\={u}}) theory, advanced by scholars like Xigui Qiu, streamlines this into: 1. \textbf{Semantographs}, whose forms relate to meaning (consolidating the first three traditional types); 2. \textbf{Loangraphs (\textit{ji\v{a} jiè})}, intralingually borrowed words for homophonic, homographic, and semantic purposes; and 3. \textbf{Phonograms (\textit{xíng sh\={e}ng})}, which combine semantic and phonetic components and form the vast majority of characters.} 

This fundamental difference means that while phonographic restoration can often be framed as a sequence modeling problem solvable by a language model \cite{assael_restoring_2022,assael_contextualizing_2025}, the ideographic nature of Chinese makes its computational study an inherently multi-dimensional, multimodal challenge\cite{zh-mo_multimodal_2024}, taking forms, sounds, meanings, and contexts in unison comprehensively. Visual information cannot be abstracted away; for tasks like rejoining, the specific graphic variants used in an inscription are crucial evidence that would be lost in a text-only representation. This makes the field less about sequence completion and more about tackling a formidable visual recognition and decipherment problem, which is compounded by several factors. Early forms of the script, for instance, exhibit significant instability, where a single word could be represented by multiple variant glyphs —-- a feature that gradually standardized over time. The character distribution follows a long-tail pattern; in the case of Oracle Bone Inscriptions, of the 3,763 characters cataloged, only 1,273 are definitively deciphered, with another 378 under debate, leaving over half (2,112) undeciphered.\footnote{Data from the Chinese Writing Museum: \url{https://www.wzbwg.com/yanjiuinfo/16170}.} The vast majority of these undeciphered characters are proper nouns, such as personal or place names, which appear infrequently. Furthermore, the script underwent significant diachronic evolution, generally moving from complexity towards simplicity and from curved to straight strokes, changes that require extensive training to trace. Finally, substantial regional variations existed, particularly during the Warring States period, when the highly diverse and chaotic scripts of the Six Kingdoms posed considerable challenges for decipherment, standing in contrast to the more standardized Qin script.

\subsection{The Dilemmas of a Golden Age}
The last few decades have been considered a golden age for paleography, marked by an explosion in the quantity of unearthed materials and a corresponding rise in the overall level of scholarly research \cite{huang_construction_2025}. However, this proliferation of data, while invaluable, has also magnified the inherent challenges of the discipline, pushing traditional, manual methods to their limits. The key difficulties are twofold, reflecting the dual challenges of data organization and research depth.

First is the \textbf{challenge of scale and collation}. Unearthed artifacts are often physically fragmented, incomplete, and poorly preserved. The scripts they bear exhibit significant stylistic, structural, and regional variations, making the collation and transcription of texts a monumental task. The sheer volume of new materials makes comprehensive analysis through purely manual means increasingly impractical. Tasks such as systematically searching for all variants of a character across thousands of artifacts or exhaustively testing hypotheses for rejoining fragments are incredibly time-consuming, if not impossible, for individual scholars or small teams.

Second is the \textbf{challenge of research depth and integration}. The decipherment of unknown or difficult characters remains a central and formidable bottleneck for the field. This process requires not only profound expertise but also the ability to synthesize vast amounts of information regarding a character's graphic form, phonological properties, semantic meaning, and contextual usage. Beyond decipherment, there is a pressing need to move beyond fragmented studies around ``new-fangled'' materials and toward more integrated, systematic research capable of revealing deeper linguistic and historical patterns. The lack of such comprehensive analysis hinders the development of a robust theoretical framework for the discipline as a whole \cite{huang_construction_2025}.

\subsection{The Computational Turn}
Faced with these dual challenges, the field has begun to embrace a new paradigm. The limitations of manual scholarship and the shortage of personnel with dual expertise in both paleography and computer science have created a clear and urgent need for technological intervention \cite{huang_construction_2025}. This has catalyzed the development of what is increasingly being called \textbf{computational paleography}.

Here, however, it is essential to clarify the scope of this position paper by distinguishing between two related but distinct concepts. Much of the work to date can be described as \textbf{paleographic digitalization}: the application of digital tools to create infrastructure, such as databases or character recognition engines. This represents a necessary and foundational stage of technological adoption. The ultimate aspiration, however, is a true \textbf{computational paleography}, a genuine scholarly discipline defined not by its tools, but by its goal: to integrate computational methods into the core logical chain of paleographical inquiry. While this field is still in its nascent stages, its principles provide a crucial framework for our analysis. This paper therefore charts the entire spectrum of progress, from foundational digitalization efforts to the first, tentative steps toward a new, computationally-informed discipline of paleography.

This turn is driven by a natural alignment between the core problems of paleography and the core capabilities of modern artificial intelligence: character images can be processed with computer vision, ancient texts can be analyzed with natural language processing, and the complex web of evidence for decipherment can be structured using knowledge graphs \cite{mo2022computational}. The goals of this computational turn can be understood as a hierarchy of increasing ambition. At the most fundamental level, computational methods serve to \textbf{augment and accelerate} traditional scholarship. They offer the potential to automate laborious tasks such as image restoration, character detection, and fragment matching, addressing the challenge of scale. More profoundly, AI can act as an \textbf{intelligent assistant} for complex reasoning. By integrating vast datasets and building knowledge graphs that connect a character's graphic form, phonological properties, semantic meaning, and usage contexts, these tools can help scholars tackle the challenge of research depth and integration. The ultimate vision is to leverage these capabilities to foster \textbf{new research paradigms}---to discover large-scale linguistic and historical patterns that are invisible to manual inspection and to innovate new methods of knowledge production for a digital era \cite{huang_construction_2025}. This survey provides a comprehensive overview of this emerging field, charting its digital resources, core methods, and future directions.

\section{Resources and Datasets}
\label{sec:resources}
The development of computational paleography has been fundamentally driven by the creation of specialized digital datasets. This section provides an overview of these critical resources, which are primarily centered on oracle bone script, reflecting its prominent role in the field, with emerging collections for other ancient scripts. We categorize these datasets based on their primary data modality and intended application, covering character-level analysis, multimodal document understanding, complex-scene processing, and specialized research tasks. Table \ref{tab:datasets} provides a comparative summary of the key datasets discussed.

\begin{table*}[b]
\centering
\caption{A summary of major datasets in computational Chinese paleography. OBI refers to Oracle Bone Inscriptions.}
\label{tab:datasets}
\resizebox{\textwidth}{!}{%
\begin{tabular}{@{}lllrrp{5cm}l@{}}
\toprule
\textbf{Category} & \textbf{Dataset} & \textbf{Script} & \textbf{\#Instances} & \textbf{\#Classes} & \textbf{Focus} & \textbf{Reference} \\
\midrule
\multirow{6}{*}{\rotatebox[origin=c]{90}{\parbox[c]{3cm}{\centering Handwritten OBI}}}
& HUST-OBS & OBI & 140,053 & 1,588 & Most comprehensive; includes deciphered and undeciphered chars. & \cite{wang_open_2024} \\
& HWOBC & OBI & 83,245 & 3,881 & Relatively balanced distribution from historical texts. & \cite{li2020hwobc} \\
& Oracle-50K & OBI & 59,081 & 2,668 & General-purpose; includes few-shot subset (Oracle-FS). & \cite{han2020accv} \\
& Ancient-3/5 & OBI & 39,009 & 1,186 & Multi-script including other early writing systems. & \cite{zhang2021ancient35} \\
& Oracle-P15K & OBI & 14,542 & 239 & Structure-aligned pairs for fine-grained analysis. & \cite{li_mitigating_2025} \\
\midrule
\multirow{4}{*}{\rotatebox[origin=c]{90}{\parbox[c]{2.5cm}{\centering Scanned OBI}}}
& OBC306 & OBI & 309,551 & 306 & Large-scale scanned images from rubbings; noisy. & \cite{OBC306} \\
& Oracle-241 & OBI & 78,565 & 241 & Mix of handwritten and scanned data. & \cite{wang2024dataset} \\
& Oracle-MNIST & OBI & 30,222 & 10 & Small, noisy dataset for baseline experiments. & \cite{wang2024dataset} \\
& OBI125 & OBI & 4,257 & 125 & Manually segmented from 1,056 rubbings. & \cite{yue2022obi125} \\
\midrule
\multirow{3}{*}{\rotatebox[origin=c]{90}{\parbox[c]{2.5cm}{\centering Other Scripts}}}
& CHUBS & Bamboo & 102,722 & -- & First large-scale dataset for Chu script. & \cite{chen_multi-modal_2024} \\
& Qin Slips Dataset & Bamboo & 66,973 & -- & First sample library for Qin bamboo slips. & \cite{chen_qin_2025} \\
& Bronze Insc. Dataset & Bronze & 120,000+ & 8,453 & Bronze character recognition. & \cite{wu_cnn-based_2022} \\
& BIRD & Bronze & 41,000+ tokens & 1,078 pairs & NLP-ready corpus for restoration \& dating; includes Glyph Net. & \cite{hua_bird_2025} \\
\midrule
\multirow{9}{*}{\rotatebox[origin=c]{90}{\parbox[c]{4cm}{\centering Multimodal \& Task-Oriented}}}
& OBIMD & OBI & 10,077 frags. & -- & Full-document multimodal annotations (bbox, transcription, etc.). & \cite{li_oracle_2024} \\
& HUSAM-SinoCDCS & Various & 2,159 images & -- & Complex scenes (calligraphy, inscriptions) for detection. & \cite{qi_ancientglyphnet_2025} \\
& EVOBC & Multi-script & 229,170 & 13,714 & Script evolution across six historical stages. & \cite{guan_open_2024} \\
& ACCID & OBI & 15,085 & 2,892 & Character- and radical-level annotations for structural analysis. & \cite{diao2023accid} \\
& OracleRC & OBI & -- & 2,005 & Radical-level decomposition for zero-shot recognition. & \cite{diao2023rzcr} \\
& RCRN & OBI & 1,606 & 362 & Noisy-clean pairs for real-world image denoising. & \cite{shi2022rcrn} \\
& PicOBI-20k & OBI & 15,175 char., 4,833 obj. & -- & Pairs OBI glyphs with real objects for visual reasoning. & \cite{chen_pictobi-20k_2025} \\
& PD-OBS & OBI & 47,157 & -- & Pairs OBI glyphs with pictographic analysis text for LVLMs. & \cite{peng_interpretable_2025} \\
& OracleSem & OBI & $\sim$26,430 & 1,762 & Glyphs with detailed descriptions and analyses for reasoning. & \cite{jiang_oraclesage_2024} \\
\bottomrule
\end{tabular}%
}
\end{table*}

\subsection{Character-Level Datasets}
\label{sec:recognition_datasets}
The majority of existing datasets consist of images of isolated characters. These resources provide a foundation for a wide range of character-level computational tasks, including recognition, component analysis, and style transfer. For oracle bone inscriptions\footnote{We only consider public OBI resources in this study. For a complete list encompassing proprietary OBI datasets and online platforms, please refer to \citet{li_comprehensive_2024}.}, they are further divided into handwritten and scanned character collections.

\paragraph{Handwritten Oracle Bone Datasets} These resources offer clean, manually traced character images, providing high-quality data for training various computational models. Representative datasets include \textbf{Oracle-50K} and its few-shot subset \textbf{Oracle-FS} (59,081 images, 2,668 classes) \cite{han2020accv}, as well as the relatively balanced \textbf{HWOBC} (83,245 samples, 3,881 classes) \cite{li2020hwobc}. The most comprehensive resource to date is \textbf{HUST-OBS}, containing 140,053 images—including 77,064 deciphered characters (1,588 classes) and 62,989 undeciphered characters (9,411 classes) \cite{wang_open_2024}. Additional specialized collections include multi-script sets like \textbf{Ancient-3} and \textbf{Ancient-5} which include other early writing systems \cite{zhang2021ancient35} and \textbf{EVOBC} which collects characters from six historical stages \cite{guan_open_2024}; and \textbf{Oracle-P15K}, a large-scale structure-aligned dataset for pairwise image analysis \cite{li_mitigating_2025}.

\paragraph{Scanned Oracle Bone Datasets} This category includes character images extracted directly from artifact rubbings or photographs, presenting more realistic challenges such as noise and image degradation. Key datasets comprise \textbf{OBC306} (309,551 instances, 306 classes) \cite{OBC306}, \textbf{OBI125} (4,257 manually segmented characters, 125 classes) \cite{yue2022obi125}, the noisy but commonly used \textbf{Oracle-MNIST} (30,222 images, 10 classes), \textbf{Oracle-241} (78,565 instances, 241 classes) \cite{wang2024dataset}, which contains both handwritten and scanned data, and \textbf{RCRN} \cite{shi2022rcrn}, designed for real-world image denoising.

\paragraph{Datasets for Other Scripts} Compared to oracle bone script, datasets for other ancient scripts are still relatively underdeveloped but are gradually expanding. For \textbf{bronze inscriptions}, a dataset of over 120,000 bronze images for character recognition was developed by \citet{wu_cnn-based_2022}; \citet{hua_bird_2025} introduced BIRD (Bronze Inscription Restoration and Dating), the first fully encoded, NLP-ready dataset comprising over 41,000 tokens, which includes a specialized Glyph Net of 1,078 grapheme-allograph pairs to support textual restoration and dating tasks. In the area of \textbf{bamboo and silk scripts}, \citet{chen_multi-modal_2024} constructed the first large-scale, open-source dataset for Chu script, CHUBS, with 164 documents, 5,033 slips, and 102,722 characters, while \citet{chen_qin_2025} assembled the first sample library for Qin bamboo slips (66,973 characters). 

\subsection{Multimodal, Full-Document, and Complex-Scene Datasets}
\label{sec:multimodal_datasets}
Recognizing the limitations of isolated character analysis, researchers have begun to develop datasets that capture artifacts in their broader context, supporting multimodal tasks and real-world challenges. The landmark resource in this area is the \textbf{Oracle Bone Inscriptions Multi-modal Dataset (OBIMD)} \cite{li_oracle_2024}, which provides comprehensive annotations for 10,077 oracle bone fragments—including bounding boxes, character classifications, transcriptions, inscription groups, and reading sequences. This enables end-to-end research on a wide range of tasks, from detection and classification to denoising, sequence prediction, and missing character completion.

To address the challenge of recognizing ancient Chinese characters in complex, real-world scenes, such as artifact photographs and especially artistic calligraphy\footnote{AncientGlyphNet and related resources go beyond Chinese paleography in the strict sense, extending their coverage into a diachronic range of calligraphy styles.}, specialized resources have been developed. \citet{qi_ancientglyphnet_2025} proposed \textbf{AncientGlyphNet}, which provides a comprehensive framework for text detection across diverse script types and historical contexts, effectively handling noisy and varied data from challenging sources. Its companion dataset, \textbf{HUSAM-SinoCDCS}, includes images from stone inscriptions, calligraphy, and couplet scenes, thereby expanding the computational scope from single-character isolation to full-scene, multimodal paleography. This approach not only covers classical ancient scripts such as oracle bone and bronze inscriptions, but also encompasses the full trajectory of Chinese character evolution—from pictographs to stylized, symbolic forms in later scripts—broadening the practical reach of computational paleography.

\subsection{Specialized and Task-Oriented Datasets}
\label{sec:task_oriented_datasets}
Beyond general recognition, a range of specialized datasets has been created to support specific paleographical research goals. These can be broadly grouped into resources for analyzing the intrinsic properties of characters, such as their historical evolution and internal structure, and those designed to power computational tasks like denoising and automated decipherment.

The first group focuses on the characters themselves. For the diachronic study of character forms, the \textbf{EVOBC} dataset provides a critical resource, offering images from six major historical scripts to enable detailed analysis of script transitions \cite{guan_open_2024}. For synchronic structural analysis, datasets with radical-level annotations facilitate a deeper understanding of character composition. \textbf{ACCID}, for instance, provides detailed annotations on radical categories, locations, and structural relations \cite{diao2023accid}, while \textbf{OracleRC} decomposes characters into their constituent radicals to support tasks like zero-shot recognition \cite{diao2023rzcr}.

The second group of datasets is tailored for specific downstream applications. For image restoration, \textbf{RCRN} offers a collection of noisy-clean image pairs for training real-world denoising models \cite{shi2022rcrn}. Pushing the frontier towards interpretation and reasoning, several resources integrate visual data with textual and semantic information. The \textbf{Radical-Pictographic Decipherment OBS Dataset (PD-OBS)} links glyphs with pictographic analysis texts for vision-language model training \cite{peng_interpretable_2025}. \textbf{OracleSem} provides glyphs with rich descriptive commentary to aid machine reasoning \cite{jiang_oraclesage_2024}, and \textbf{PicOBI-20k} pairs characters with real-world object images to benchmark the visual reasoning capabilities of multimodal models \cite{chen_pictobi-20k_2025}.

\subsection{Challenges in Data Curation}
\label{sec:data_challenges}
Despite the growing number of resources, the creation of high-quality, comprehensive datasets for computational paleography is fraught with challenges that significantly impact the field's development.
\begin{enumerate}
    \item \textbf{Severe Distributional Skew:} A major imbalance exists in the available data. Research and dataset collection have disproportionately focused on oracle bone script, where data is most abundant, leaving bronze inscriptions and bamboo/silk scripts comparatively under-resourced. This skew creates significant obstacles for diachronic studies aiming to trace the evolution of characters across different historical periods and media.
    \item \textbf{Inconsistent Data Quality:} The quality of existing datasets varies widely. They are often aggregated from a mix of public and private sources with differing curation standards. Many suffer from severe class imbalance, reflecting the long-tail distribution of characters in ancient scripts. Furthermore, the nature of the source materials introduces challenges like physical fragmentation, surface damage, and noise from rubbings or photographs.
    \item \textbf{Lack of Standardization:} There is a notable absence of a unified framework for dataset creation and documentation \cite{huang_construction_2025}. This lack of systematic organization leads to duplicated efforts, hinders the reproducibility and reusability of datasets, and risks wasting resources on projects that are not sufficiently grounded in paleographical standards. Basic infrastructure is also uneven --- many resources are not publicly available or are scattered across various online platforms, often requiring laborious manual curation. The same situation is reflected in data construction for computational paleography, where much data remains inaccessible and non-standardized, with researchers sometimes needing to collect and prepare resources themselves from disparate sources.
    \item \textbf{Modality Imbalance:} Existing datasets are heavily skewed towards isolated character glyphs. Resources representing characters in their natural context, such as on full artifact rubbings or photographs, remain scarce. Moreover, there is a significant lack of data capturing the textual and documentary context essential for semantic analysis. While phonetic data could offer another valuable dimension, its collection is hampered by the profound difficulty of accurately reconstructing ancient Chinese phonology.
    \item \textbf{Bridging the Semantic Gap:} Current research predominantly focuses on low-level tasks such as character recognition and decipherment. There is a significant semantic gap between these tasks and the higher-level interpretive work that is the hallmark of paleography, such as tracing scribal hands, dating manuscripts, and understanding the evolution of letterforms in their historical context.
    \item \textbf{Challenges in Expert-Informed Annotation:} While many studies adopt a paradigm of AI-assisted annotation followed by expert review, significant challenges remain. A core issue is the quality and nature of the annotated data. There is a scarcity of datasets that are not only labeled by experts but are also designed to address perspectives genuinely valuable to paleographical inquiry, especially for higher-level tasks like interpretation and semantic analysis. 
\end{enumerate}

\section{Foundational Visual Analysis}
\label{sec:visual_analysis}
The first stage in the computational paleography pipeline involves processing raw images of artifacts to restore, enhance, and ultimately recognize the characters they contain. This section covers the core visual analysis tasks, from preprocessing and data augmentation to the fundamental challenge of character recognition.

\subsection{Image Processing and Augmentation}
\label{sec:process}
A critical first step in analyzing artifact images is addressing the pervasive issues of noise, physical damage, and low contrast. Research in this area focuses on denoising, data augmentation, and inpainting. To handle noise from rubbings and photographs, early methods relied on traditional filters like K-SVD and anisotropic diffusion \cite{shi2017chinese, huang2016comparison}, while recent work has shifted to deep learning. \citet{wang_unsupervised_2022} introduced a Structure-Texture Separation Network (STSN) to disentangle character strokes from background noise. For oracle bone inscriptions, \citet{li_obiformer_2025} developed OBIFormer, a computationally efficient framework that leverages channel-wise self-attention and selective kernel feature fusion for precise image reconstruction. To better handle complex character structures, another line of work proposes incorporating standard character writing model (tianzige), using local branches to supplement global features alongside a refined loss function to prevent strokes from adhering to noise \cite{miao_research_2022}. For the common problem of data scarcity, generative models have emerged as a powerful solution for \textbf{data augmentation}. Beyond general-purpose generation using GANs \cite{huang_agtgan_2023}, controllable models like Diff-Oracle have been introduced to generate diverse and high-quality oracle characters \cite{li_diff-oracle_2024}. Diff-Oracle utilizes separate style and content encoders, allowing it to synthesize new characters by combining stylistic features and glyph information from different reference images. Other tailored strategies have also been developed, including methods specifically designed for long-tailed distributions, such as cutting and pasting patches from rare to frequent classes (Repatch) or adapting mixing ratios based on class priors (TailMix) \cite{li2023towards}, and context-aware techniques that isolate the character foreground before applying transformations \cite{zhang2021bag}. Another novel approach involves rasterizing glyphs into stroke vectors and using language models like BERT to augment the stroke sequences before re-rendering them as images \cite{han2020accv}. For physically damaged artifacts, \textbf{image inpainting} has also seen significant advances. The Multi-Modal Multitask Restoration Model (MMRM) represents a pioneering approach that synergizes textual and visual information. It combines context understanding with residual visual data from damaged artifacts to simultaneously predict missing characters and generate restored images \cite{duan_restoring_2024}.

Once images are cleaned and augmented, the core task is to 1) detect and 2) recognize individual characters. This has evolved from classical feature-based methods to modern deep learning architectures.

\subsection{Character Detection and Recognition}
\label{sec:reg}

\paragraph{Classical Methods}
Early approaches relied on handcrafted structural and statistical features. For instance, some methods treated characters as graphs and used topological properties for recognition \cite{zhou1995jia, li2011isomorphism}, while others used techniques like the Hough Transform for structural analysis \cite{meng2017recognition}. Scale-Invariant Feature Transform (SIFT) has been applied for robust keypoint extraction in damaged inscriptions \cite{sun2020dual}, and various encoding schemes have been used to create compact representations for efficient matching \cite{chen2020study}. Other methods have focused on specific feature types, such as Histogram of Oriented Gradients (HOG) features combined with a Support Vector Machine (SVM) classifier \cite{ou2024qin}. For scripts with complex structures like small seal script, structural hashing methods that leverage the spatial relationships between character pairs have been employed to achieve rotation and scale-invariant recognition \cite{roy2009seal}. More recently, \citet{tao_clustering-based_2025} proposed a clustering-based feature representation learning method that uses glyph instances as prior knowledge to improve feature extraction in detection networks.

\paragraph{Deep Learning Methods}
The field is now dominated by deep learning. Early work, such as the two-stage feature mapping network by \citet{li_deep_2018} for bronze inscriptions, demonstrated the power of Convolutional Neural Networks (CNNs). Modern approaches have since incorporated more advanced architectures and techniques. Standard CNNs have been enhanced with Inception modules, residual connections, and attention mechanisms \cite{mai_oracle_2024, guo2022improved, luo2023aggregation}. A notable trend is the development of methods grounded in the logographic nature of Chinese characters, which treat characters as compositions of radicals. These \textbf{radical-based methods} first learn to identify constituent components before recognizing the whole character, which can improve performance on complex characters and support zero-shot learning \cite{luxuzheng2020, lin2022radical}. The Transformer architecture has also been influential; \citet{wu2021ancient} used a Transformer-based model with multi-scale attention for Chu bamboo slip characters, and \citet{chen_qin_2025} proposed the lightweight QBSC Transformer, which fuses separable convolutions with windowed self-attention to achieve high accuracy on Qin slips with minimal parameters. Beyond direct recognition, deep learning features have been used for related analytical tasks, such as combining ResNet50 features with spectral clustering to automatically group character variants \cite{liu_recognition_2021} or using prior knowledge to distinguish oracle variants based on isomorphism and symmetry \cite{WOS000564295200001}.

\paragraph{Advanced Learning Strategies}
To further push performance boundaries, especially given the challenges of data scarcity and imbalance, researchers have explored a range of advanced strategies. To tackle the long-tail distribution of characters, \textbf{imbalanced learning} techniques such as deep metric learning with triplet loss \cite{icdar2019, zhang2019oracle} and decoupled training strategies \cite{li2023decouple} have been employed. For \textbf{few-shot or zero-shot recognition}, methods have focused on data augmentation through self-supervised learning on modern characters \cite{han2020accv} or by leveraging radical-based reasoning to recognize unseen characters. These approaches often rely on external knowledge; for example, by learning style-independent radical sequences \cite{zhou2023style} or by traversing an explicit Character Knowledge Graph with a graph convolutional network to infer relationships between characters and radicals \cite{diao2023rzcr}. Another key strategy is \textbf{cross-modal learning} or unsupervised domain adaptation, which aims to leverage clean, handwritten character data to improve recognition on noisy, scanned images. The Structure-Texture Separation Network (STSN), for instance, was designed to disentangle a character's core glyph structure from background noise, enabling more robust adaptation \cite{wang_unsupervised_2022}. Finally, \textbf{ensemble learning} has also shown promise, with systems combining multiple CNNs \cite{chen_recognition_2025} or a mix of ViT and ResNet models \cite{wang_innovative_2025} to boost performance.

Despite these advanced strategies, the task remains formidable due to several intrinsic difficulties. A core issue is the immense \textbf{stylistic diversity} and the prevalence of character variants. This challenge is twofold: beyond simple allographic variants arising from different scribal hands, ancient scripts feature a vast number of structural variants where the fundamental composition of a character can differ, a much harder problem than modern character recognition. Moreover, the available data is plagued by \textbf{sparsity and a highly skewed distribution}. Deep learning models are data-driven, yet for ancient scripts, many characters appear only once or twice in the entire corpus, providing insufficient data for models to learn robust features and generalize effectively. The visual quality of the source material presents another hurdle, as images from rubbings or photographs are frequently marred by \textbf{signal degradation}, including physical cracks, background noise, and partial occlusion of strokes. The preliminary task of \textbf{character localization} often proves to be a critical bottleneck; this essential prerequisite for recognition is itself a challenging and relatively underexplored problem, especially on crowded inscriptions or degraded surfaces where its accuracy dictates the success of the entire workflow. Finally, real-world application demands a solution to the \textbf{open-set conundrum}, where systems must not only classify known characters but also gracefully handle novel or undeciphered glyphs, a capacity that is crucial for any practical paleographical tool.

\section{Contextual Analysis: From Artifacts to Texts}
\label{sec:contextual_analysis}
\subsection{Artifact Rejoining and Deduplication}
\label{sec:rejoining}
A significant challenge in paleography is the physical \textbf{rejoining}/reconstruction (\textit{zhuì hé}) of fragmented artifacts, such as oracle bones and bamboo slips, which have often been buried for millennia and suffered severe breakage. The process of rejoining these fragments is traditionally a painstaking and time-consuming manual task for experts. Computational methods have been developed to automate and accelerate this work by analyzing a combination of visual and textual features, such as the physical contours of fracture edges and the sequence of inscribed characters. For oracle bones, for example, approaches range from Siamese networks for fragment matching \cite{WOS001102506600004} to the SUM algorithm, which uses contour slope and character sequences \cite{zhang_oracle_2023}. For bamboo slips, the WisePanda framework uses a physics-driven deep learning model to significantly improve matching accuracy and efficiency \cite{zhu_rejoining_2025}.

A related challenge is \textbf{deduplication} (\textit{jiào chóng}), the process of identifying multiple rubbings of the same artifact, which often exist due to historical collection and circulation. This foundational collation task has been significantly advanced by computational methods. AI models approach this problem by operating on the raw image, treating the inscription as a visual pattern rather than as text. This allows them to identify duplicates in challenging cases where the textual content is not perfectly identical, which are often of high scholarly value \cite{zh-mo_application__2023}. Computational work in this area includes building benchmarks for homologous rubbing retrieval \cite{WOS001514394500002} and developing frameworks like OBD-Finder, which combines keypoint matching with textual analysis to successfully discover new duplicate fragments \cite{zhang_explainable_2025}.

\subsection{Periodization and Classification}
\label{sec:dating}
Determining the age and origin of artifacts, or \textbf{periodization} (\textit{duàn dài}) is a core task in historical analysis. For oracle bone inscriptions, font classification serves as a crucial basis for periodization, which has been tackled with recurrent graph neural networks \cite{WOS001150626700002}. For character variant analysis, \citet{liu_recognition_2021} combined ResNet50 feature extraction with spectral clustering to automatically group visually similar glyph variants, aiding in the study of script evolution and regional styles.

A key application of this technology is the automated dating of bronze vessels, a task traditionally requiring years of expert training. By building intelligent models, this high barrier to entry can be significantly lowered. A representative study by \citet{zh-li_ding__2023} demonstrates a comprehensive approach to dating ancient bronze \textit{d\v{i}ng} tripods. They constructed a dataset of over 3,600 vessel images, annotating them not only by period but also with fine-grained typological details such as 29 shape variations and 67 decorative pattern types. Their deep learning model, which uses a knowledge-guided graph to connect these features, achieved 78.8\% accuracy on a fine-grained, 11-period classification task and 89.8\% on a coarse, 4-dynasty task. Their error analysis revealed a crucial insight: like human experts, the model's misclassifications occurred most frequently between adjacent historical periods, where stylistic evolution was gradual. Beyond visual features, \citet{hua_bird_2025} demonstrated that textual content also carries strong chronological signals; their BIRD framework employs glyph-biased sampling to emphasize historically informative allographs, achieving competitive accuracy in dating bronze inscriptions solely from text. 

\subsection{Segmentation and Language Modeling}
\label{sec:nlp}
Analyzing inscriptions as language requires segmenting character sequences and building language models. To accelerate the deciphering of oracle bone script, \citet{hu_component-level_nodate} proposed the new task of component-level segmentation, inspired by a successful 2018 deciphering case. Their model leverages both annotated and weakly annotated data along with expert-defined stroke rules to help specialists quickly identify character components. For bamboo slips, where ink bleed and background noise are common, \citet{cao_character_2022} introduced a local adaptive thresholding method that uses an effective character contour length metric and multi-Gaussian fitting to determine the optimal segmentation threshold. Bridging the gap between ancient scripts and modern NLP, \citet{chen_multi-modal_2024} designed a multi-modal, multi-granularity tokenizer for Chu bamboo scripts that maps glyphs to modern characters or sub-character units. Alongside this, they assembled the first large-scale dataset of Chu slips with over 100,000 annotated characters, and their tokenizer achieved a 5.5\% relative F1-score improvement on part-of-speech tagging. To build language models for ancient Chinese, \citet{diao_intechar_2025} created InteChar, a unified character list integrating un-encoded oracle bone characters with modern Chinese. They also constructed the Oracle Corpus Set (OracleCS), combining expert annotations with LLM-assisted data augmentation, providing a foundational resource for training more capable models. Similarly, for bronze inscriptions, the BIRD framework formulates restoration as an allograph-aware masked language modeling problem \cite{hua_bird_2025}. By integrating a Glyph Net that links graphemes with their allographs, the model can effectively leverage allographic similarity to restore missing characters in fragmented texts, stabilizing predictions even in a low-resource setting.

\section{Advanced Reasoning and Knowledge Systems}
\label{sec:advanced_reasoning}
The frontier of computational paleography lies in developing systems that can perform complex reasoning, assist in the core intellectual task of decipherment, and integrate vast repositories of knowledge into accessible, human-centric tools.
 
\subsection{Automated Decipherment}
\label{sec:decipherment}
Automated decipherment, the ultimate goal of ancient character recognition, represents the most innovative and challenging frontier of computational paleography. Here, we define decipherment in its strict paleographical sense: not as synonymous with character recognition, but as the scholarly process of interpreting undeciphered characters (\textit{k\v{a}o shì}) and elucidating the principles of their original construction (\textit{zào zì l\v{i} jù}). There is a scholarly consensus that this task is well-suited for artificial intelligence; because decipherment is a problem with clear boundaries that should, ``in principle, have a single correct solution, it is an ideal candidate for AI-assisted research'' \cite{wu_paleography_2020}. Crucially, the methodological foundation for computational approaches remains grounded in the traditional methods of paleographical experts. The classical methods of decipherment, such as comparison of graphic forms, contextual deduction, component/radical analysis, and historical verification, ``find direct parallels in modern AI techniques''. Image recognition maps to form comparison, natural language processing to contextual analysis, and knowledge graphs to historical verification \cite{li_how_2022}. The following sections review the emerging computational approaches, which can be broadly categorized in alignment with these classical principles.

\subsubsection{Generative and Diffusion-Based Methods} 

One promising direction involves leveraging generative models to produce visual and textual clues for unknown characters. A landmark example is OBSD \cite{guan_deciphering_2025}, which addresses the significant structural gap between ancient and modern characters via a two-stage process: first, a conditional diffusion model with a \textit{Localized Structural Sampling} strategy generates an initial form from image patches; second, a \textit{zero-shot refinement} module, trained on various modern fonts, corrects structural artifacts. The model directly generates modern character glyphs from OBI images, achieving plausible results on deciphered characters and producing visual hints for undeciphered ones; however, these hints are often of limited practical value to paleographical experts. A similar approach, DCSD-OBI, attempts to improve semantic integration by using a dual-conditional stable diffusion model that incorporates both OBI images and modern Chinese text, further enhancing performance with efficient fine-tuning and a specialized Chinese-CLIP model \cite{WOS001570216400003}. Despite these advances in semantic alignment, its utility for experts in the novel decipherment of unknown characters remains similarly limited.

Another approach, OracleFusion\cite{li_oraclefusion_2025}, aims to assist experts by generating an interpretable ``semantic typography''—a semantically rich vectorized font—rather than a direct translation. It employs a two-stage process: first, a Multimodal Large Language Model (MLLM) analyzes the ancient glyph to produce a textual description of its components and their spatial layout. Second, this structured description guides a vector generation model to create a clean, stylized rendering of the oracle character that emphasizes its semantic parts. The stated goal is to provide a visually enhanced and structurally explicit representation to aid human interpretation. However, this approach is highly dependent on the model's prior knowledge, and the generated glyph interpretations for already-deciphered characters can often be paleographically unsound, making them more of a technical demonstration than a reliable scholarly tool. 

\subsubsection{Compositional and Component-Based Methods} 

Grounded in the logographic structure of Chinese characters, these methods treat decipherment as a compositional task. The Puzzle Pieces Picker (P$^3$) framework, for instance, treats decipherment as a ``jigsaw puzzle'' problem, where  characters are deconstructed into components and then reconstructed into their modern equivalents \cite{wang_puzzle_2024}. To achieve this, it first creates a radical-level dataset through an automated pipeline: character images are segmented into potential radicals using both contour detection and the Segment Anything Model (SAM), and a self-supervised model learns feature representations for these components, which are then clustered and labeled via KNN. The core objective is to take an ancient character image as input and output a reconstruction recipe: a sequence of modern radicals and an Ideographic Description Sequence (IDS) code that specifies their spatial arrangement (e.g., left-right, top-bottom). While validating P$^3$'s effectiveness, the study also demonstrated that incorporating data from historically adjacent periods, such as Bronze script, significantly boosted performance, confirming the model's ability to leverage evolutionary context. 

Similarly, the CoLa is inspired by the human ability to recognize new characters by deconstructing them into familiar parts \cite{shi_cola_2025}. Instead of relying on predefined radical systems, its central idea is to have the model learn its own visual vocabulary of character components directly from images. It achieves this by forcing the model to represent characters as a small collection of abstract ``latent components,'' which must then be recombined to reconstruct the original character's high-level visual features. This process compels the model to discover a meaningful and efficient decomposition strategy on its own, without human supervision. Recognition of unknown characters is then achieved by comparing their automatically discovered component signatures. This approach also demonstrates generalization; for instance, when trained only on later historical scripts, CoLa can successfully parse the structure of unseen Oracle Bone Characters. 

Other work explicitly uses Large Vision-Language Models (LVLMs) to mimic the interpretive process of human experts \cite{peng_interpretable_2025}. The core idea is that decipherment requires bridging the gap between a character's visual form and its semantic meaning. To achieve this, the LVLM is taught to perform a dual analysis: first, identifying a character's primary radical and analyzing its semantic contribution, and second, interpreting the pictographic meaning of the character's overall shape. This is enabled by a progressive training strategy on a new, purpose-built dataset (PD-OBS) containing expert-style textual analyses. For the final decipherment, instead of direct prediction, the model generates textual descriptions from its dual analysis. These descriptions are then used to retrieve the most semantically similar modern characters from a comprehensive dictionary. This method not only achieves strong zero-shot performance but also provides an interpretable, step-by-step reasoning chain that can offer valuable insights for undeciphered characters.

\subsubsection{Cross-Reference and Multi-Modal Reasoning} 

These approaches seek to decipher characters by integrating multiple sources of information. 

One approach to decipherment, inspired by traditional paleographic practices, is cross-font image retrieval \cite{wu_cross-font_2024}. The core idea is that deciphering an unknown OBI often involves comparing it with characters from other, historically related script forms to find evolutionary links. This method proposes a Cross-Font Image Retrieval Network (CFIRN), built on a siamese framework, to automate this comparative analysis. The network learns to extract robust, deep features from character images across various fonts, using a Multiscale Feature Integration (MFI). Given an undeciphered OBI, the CFIRN retrieves its most similar counterparts from a gallery of characters in different historical scripts, facilitating decipherment by providing plausible intermediary character forms and thereby bridging the significant historical gaps in character evolution.

Extending the idea of retrieval-based decipherment, the LUC (Linking Unknown Characters) framework focuses on retrieving similar oracle bone characters directly from raw rubbing images, rather than just clear glyphs \cite{WOS001204747000001}. The core challenge it addresses is bridging the ``domain gap'' between clean input glyphs (e.g., handwritten by users or downloaded) and noisy, real-world rubbing images. This is achieved through a deep learning retrieval framework that employs a domain-aware embedding module. This module generates oracle bone radical prototypes to enhance the structural features extracted from the characters, making the comparison between clean glyphs and rubbing images more effective.

While most decipherment work focuses on visual features, contextual analysis is also being explored. In one study, \citet{zh-mo_initial_2021} trained a BERT model on a large classical Chinese corpus to predict masked characters in Warring States bamboo slips. Out of a vocabulary of over 23,000 characters, the model placed the correct character in the top 5 predictions 20\% of the time and in the top 10 25\% of the time. It is concluded that such data-driven methods simulate the experience-based component of decipherment but must be combined with knowledge-based approaches to replicate expert reasoning.

The OracleSage framework, on the other hand, introduces a novel cross-modal approach to ancient text interpretation that explicitly unifies visual and linguistic understanding, closely mirroring how human experts decipher Oracle Bone Scripts \cite{jiang_oraclesage_2024}. Its core strategy is to interpret characters from a dual perspective: through their visual morphology and their embedded semantics. This is achieved via two main components: a Hierarchical Visual-Semantic Understanding module, which progressively fine-tunes an LVLM's visual backbone to extract multi-granularity features; and a Graph-based Semantic Reasoning, which models the complex relationships between these visual components, their structural arrangements, and abstract semantic concepts. This framework uses dynamic message passing on a heterogeneous graph to reason about meaning.

\subsubsection{Benchmarks}

The rapid development of these diverse strategies has underscored the need for systematic evaluation. Recent benchmarks like OBI-Bench have assessed the performance of large multimodal models across the full pipeline (recognition, rejoining, classification, retrieval, and deciphering) of oracle bone tasks, revealing that current models barely reach the level of an untrained human \cite{chen_obi-bench_2025}. Reasoning frameworks like V-Oracle have begun to frame decipherment as a visual question-answering problem, using a multi-dimensional, multi-step chain of reasoning and multi-stage alignment to achieve significant improvements on custom decipherment and interpretation tasks \cite{qiao_v-oracle_2025}. These efforts are beginning to systematically assess the capabilities of large models on these complex decipherment tasks, pushing the field towards more rigorous and paleographically-grounded evaluation.

\subsection{Human-AI Collaborative Systems}
\label{sec:collaboration}
Recognizing the irreplaceable role of expert intuition, the field is increasingly moving towards human-AI collaborative systems that augment, rather than replace, human expertise. This paradigm is built on the principle of methodological synergy: combining the high-level cognitive and contextual reasoning of human scholars with the ``brute-force'' capacity of AI for low-level feature analysis. 

This collaborative approach has been highly influential. Two milestone papers from DeepMind on ancient Greek epigraphy demonstrated the power of this synergy. The Ithaca system for textual restoration, geographical attribution, and dating raised the accuracy of historians from 25\% to 72\% in a collaborative setting \cite{assael_restoring_2022}. Its successor, Aeneas, further enhances this potential by integrating visual and textual information with a parallel retrieval mechanism and a progressive restoration strategy \cite{assael_contextualizing_2025}.

In Chinese paleography, this principle is particularly evident in the task of artifact rejoining. Although the ambition to use computers for this task dates back to the 1970s, it is only through recent advances in computational resources that this vision is gradually becoming a practical reality\cite{zh-wang_computer_2011,zh-wang_intelligent_2010,zh-wang_computer_2010-2}, leading to new discoveries \cite{zh-mo_computer_2021-1}. While human experts traditionally rely on methods requiring deep scholarly knowledge, such as analyzing font styles, remnant characters, and parallel texts, the matching of physical fracture edges (\textit{chá k\v{o}u}) represented a historical blind spot, used more for verification than for discovery due to the cognitive difficulty of matching countless shapes at scale \cite{zh-mo_application__2023}. This exemplifies the core insight of ``human-AI coupling'': delegating computationally intensive but conceptually straightforward (``simple nature'') tasks to AI, while humans focus on complex, integrative problems that require creativity and accumulated knowledge (``complex nature''). \citet{zh-mo_rejoining_2021} provides a compelling case for this synergy, demonstrating that by having scholars handle nuanced font-style analysis while AI manages large-scale fracture-edge matching, successful new rejoinings have been achieved, particularly for fragments lacking any textual overlap. This principle is also embodied in systems like RejoinX (``\textit{zh\={i} w\={e}i  zhùi},'') which uses deliberate visualization stimulus to guide expert intuition to successfully rejoining of new oracle bone fragments \cite{li_aiguided_2024}, and GenOV\cite{qiao2024making}, featuring a large vision-language model for reasoning with a text-to-image model to generate visual guides that help make the ancient script more accessible to the public.

\subsection{Knowledge Graphs and Document Collation}
\label{sec:kg}
To manage the vast and interconnected web of paleographical knowledge, researchers are building knowledge graphs and intelligent collation systems. To solve issues of knowledge fragmentation and disconnection, for example, a multi-modal knowledge graph for oracle bone studies has been proposed to provide a unified semantic space for heterogeneous data \cite{WOS000663706800002}. \citet{li_ancient_2023} discussed the architecture of a pyramid knowledge graph, encompassing form, radical, character, word, sememe, and vocabulary, and its use in the quantitative study of similar characters. In a practical application of graph theory, \citet{li_diviner_2021} modeled the social networks of diviners in Shang dynasty oracle bone inscriptions. From an analysis of 61,043 oracle bone fragments, they identified 828 instances of co-occurrence involving 55 diviners and 139 relationship pairs distributed across four groups. These connections were then visualized as a network graph and an adjacency matrix heatmap. For document collation, the LeverX (``\textit{zh\={i} diǎn}'') system integrates text-image multimodal information to support new annotation schemas targeting key challenges in oracle bone studies, including Named Entity Annotation, Divinatory Focus Annotation, Inscription Layout Visualization, Inscription-Crack Graphs, and Inscription Mapping Graphs. By streamlining the research process and reducing repetitive labor, the system enables scholars to focus on deepening content, improving quality, and expanding academic insights \cite{li_new_2024}.

\section{Discussion: Challenges and Future Directions}
\label{sec:discussion}
While computational paleography has made significant strides, it remains a nascent field facing substantial challenges and poised at the intersection of several key developmental trends.

\subsection{Challenges}
\label{sec:challenges}
The foremost challenge is not technical but human: fostering a deep and effective spirit of interdisciplinary collaboration. The success or failure of this entire enterprise will be determined by the depth of integration between computational science and paleographical scholarship. Any meaningful progress requires mutual learning: computer scientists must take the initiative to understand the nuances of paleography, just as young paleographers have begun to engage with AI and programming\cite{mo2022computational}. This is the critical insight, for computers are not omnipotent. AI still faces profound limitations in causal reasoning, inference, and true understanding. The technical obstacles listed below are significant, but they can only be overcome through a genuine partnership where technology serves, rather than dictates, the course of humanistic inquiry.

Alongside this overarching need for collaboration, the field faces several core technical and data-related obstacles:
\begin{enumerate}
    \item \textbf{Data Scarcity and Quality:} Ancient script samples are inherently limited. Their distribution is long-tailed, with a few common characters and thousands of rare ones. The data is also of poor quality, suffering from physical fragmentation, textual corruption, and significant noise in images of rubbings and artifacts. Critically, many existing datasets do not reflect the true distributional characteristics of historical corpora. They are often constructed by intentionally selecting balanced or artificially diversified character sets, which may yield favorable benchmark results but limit the real-world applicability of the resulting models.
    \item \textbf{Uneven Research Focus:} Research has disproportionately focused on oracle bone script, for which data is most abundant, leaving bronze inscriptions, bamboo slips, and other scripts comparatively under-resourced.
    \item \textbf{Lack of Semantic Understanding and Modality Imbalance:} The logographic nature of Chinese script makes multimodality essential, yet current models remain unbalanced. Image-based models are far more mature, while language models for ancient texts lag significantly. This is for two main reasons. First, ancient language corpora are too sparse to train large models from scratch, necessitating a transfer-learning approach from the larger corpus of received/trainsmitted texts to excavated ones. Second, core paleographical tasks are more visually entangled than in phonographic systems; decipherment, for instance, is a complex task of linking ancient and modern glyphs that barely exists for alphabetic scripts, while artifact rejoining depends heavily on visual cues from character variants that are lost in text-only representations. This forces most systems to operate in a single modality, acting as "assistants" that require human experts to provide the missing linguistic or visual context to complete a task \cite{zh-mo_multimodal_2024}.
    \item \textbf{Generalization Across Domains:} Models trained on one script type, time period, or medium (e.g., bronze vs. bamboo) often fail to generalize to others due to significant variations in style and form.
\end{enumerate}

Perhaps most critically, there is a severe scarcity of expert perspectives that are genuinely valuable to paleographical inquiry. Peer-reviewed datasets designed for higher-level tasks like interpretation and semantic analysis are particularly rare. This creates an underlying risk that AI research, while technically impressive, may operate in a vacuum and fail to address the core questions of paleography. Without the deep involvement of domain experts, computational work can become a self-referential exercise with little meaningful impact on the humanities discipline it aims to serve.

\subsection{Future Directions}
\label{sec:future_directions}
The field is moving towards more integrated and intelligent systems, characterized by several major trends that reflect a deepening of the partnership between AI and humanistic inquiry \cite{zh-mo_application__2023}:
\begin{enumerate}
    \item \textbf{From Visual to Unified Multimodal Models:} Ancient script is inherently multimodal, existing as both image and text. Future work must move beyond processing these as separate streams and toward unified models that can reason across visual, textual, and structured knowledge simultaneously with well-adjusted agency in tasks at all scales in every step. This reflects a more holistic approach that mirrors the synthetic reasoning of human experts.
    \item \textbf{From Automation to Human-AI Synergy:} The goal is shifting from replacing human tasks to augmenting human intellect. This includes building interactive systems that act as intelligent assistants and, more profoundly, using generative AI for \textit{divergent thinking}. In this paradigm, AI serves as an intellectual partner, providing inspiration and novel perspectives to help scholars overcome cognitive bottlenecks, to ``pierce the window paper'' of a difficult research problem.
    \item \textbf{From Task Execution to Knowledge Discovery:} Beyond solving discrete tasks like recognition or rejoining, a key future direction lies in using AI for large-scale data mining. By analyzing vast corpora of paleographical materials, AI can uncover novel, large-scale linguistic and historical patterns that are invisible to manual inspection, thereby generating new theoretical insights for the discipline.
    \item \textbf{From Supervised to Few-Shot and Zero-Shot Learning:} Given the inherent scarcity of data for ancient scripts, the field will continue to move away from data-hungry supervised methods and toward models that can learn effectively from limited examples and generalize to unseen character classes, which is essential for handling rare or undeciphered scripts.
\end{enumerate}
Ultimately, the future of the field hinges on a problem-oriented approach that is deeply grounded in the traditions and knowledge of classical paleography. Most current AI research remains focused on the ``form'' and ``meaning'' of characters, while neglecting the crucial dimensions of ``sound'' and ``contextual evidence''. Given the immense challenge of reconstructing ancient phonology, a promising path forward lies in leveraging computational methods to build a new generation of scholarly infrastructure for textual analysis. This includes developing more semantically-informed image processing tools, such as structure-aware denoising algorithms that can distinguish between genuine strokes and stroke-like noise, alongside tools for fuzzy search, intelligent document analysis, flexible knowledge retrieval, and open question answering to empower the next wave of paleographical research.

\section{Conclusion}
\label{sec:conclusion}
This paper has charted the rapid emergence of computational Chinese paleography, documenting its evolution from the automation of discrete visual tasks to the creation of integrated research platforms. Our analysis reveals a field in transition, moving beyond a narrow focus on character recognition toward a more holistic vision of multimodal systems designed to assist in complex scholarly inquiry. We have systematically traced this progression, from foundational image analysis and contextual reconstruction to the frontier of automated decipherment and knowledge discovery.

Despite this progress, a central tension remains: the disjunction between the data-intensive nature of modern AI and the sparse, fragmented, and noisy reality of ancient artifacts. The path forward, as we have argued, does not lie in the pursuit of full automation but in fostering a deeper synergy between human expertise and machine intelligence. The most promising research will prioritize the development of human-AI collaborative systems, leveraging few-shot learning and multimodal data fusion. The ultimate goal is to build a new generation of intelligent scholarly infrastructure that augments, rather than replaces, the interpretive skills of paleographers, thereby unlocking new frontiers in our understanding of ancient civilization.

\begin{acknowledgments}
    We are indebted to Prof. Shuangjie Li, together with the Research and Conservation Center for Unearthed Texts, Tsinghua University. Their invaluable guidance and pioneering vision shaped this work, instilling in us a profound reverence for Chinese paleography and an immense hope for its computational future. We sincerely wish this nascent field a brilliant journey ahead.
\end{acknowledgments}


\bibliographystyle{compling}
\bibliography{pal}

@article{assael_contextualizing_2025,
	author = {Assael, Yannis and Sommerschield, Thea and Cooley, Alison and Shillingford, Brendan and Pavlopoulos, John and Suresh, Priyanka and Herms, Bailey and Grayston, Justin and Maynard, Benjamin and Dietrich, Nicholas and Wulgaert, Robbe and Prag, Jonathan and Mullen, Alex and Mohamed, Shakir},
	doi = {10.1038/s41586-025-09292-5},
	issn = {0028-0836, 1476-4687},
	journal = {Nature},
	language = {en},
	month = {September},
	number = {8079},
	pages = {141--147},
	title = {Contextualizing ancient texts with generative neural networks},
	url = {https://www.nature.com/articles/s41586-025-09292-5},
	urldate = {2025-09-29},
	volume = {645},
	year = {2025}
}

@article{assael_restoring_2022,
	author = {Assael, Yannis and Sommerschield, Thea and Shillingford, Brendan and Bordbar, Mahsa and Pavlopoulos, John and Chatzipanagiotou, Marita and Androutsopoulos, Ion and Prag, Jonathan and De Freitas, Nando},
	doi = {10.1038/s41586-022-04448-z},
	journal = {Nature},
	number = {7900},
	pages = {280--283},
	title = {Restoring and attributing ancient texts using deep neural networks},
	volume = {603},
	year = {2022}
}

@article{cao_character_2022,
	author = {Cao, Songxiao and Shu, Zichao and Xu, Zhipeng and Xie, Dailiang and Xu, Ya},
	doi = {10.1007/s11042-022-11988-z},
	issn = {1380-7501, 1573-7721},
	journal = {Multimedia Tools and Applications},
	language = {en},
	month = {March},
	number = {6},
	pages = {8199--8213},
	title = {Character segmentation and restoration of {Qin}-{Han} bamboo slips using local auto-focus thresholding method},
	url = {https://link.springer.com/10.1007/s11042-022-11988-z},
	urldate = {2025-10-02},
	volume = {81},
	year = {2022}
}

@article{chen2020study,
	author = {Chen, Tingzhu and Qian, Yaoyao and Pei, Jingyu and Wu, Shaoteng and Wu, Jiang and Li, Lin and Tu, Jung-yueh},
	journal = {Journal of Chinese Writing Systems},
	number = {4},
	pages = {281--290},
	publisher = {SAGE Publications Sage UK: London, England},
	title = {A study on encoding-based oracle bone script recognition},
	volume = {4},
	year = {2020}
}

@misc{chen_multi-modal_2024,
	author = {Chen, Yingfa and Hu, Chenlong and Feng, Cong and Song, Chenyang and Yu, Shi and Han, Xu and Liu, Zhiyuan and Sun, Maosong},
	doi = {10.48550/arXiv.2409.01011},
	month = {September},
	note = {arXiv:2409.01011 [cs]},
	publisher = {arXiv},
	title = {Multi-{Modal} {Multi}-{Granularity} {Tokenizer} for {Chu} {Bamboo} {Slip} {Scripts}},
	url = {http://arxiv.org/abs/2409.01011},
	urldate = {2025-09-29},
	year = {2024}
}

@misc{chen_obi-bench_2025,
	author = {Chen, Zijian and Chen, Tingzhu and Zhang, Wenjun and Zhai, Guangtao},
	doi = {10.48550/arXiv.2412.01175},
	month = {February},
	note = {arXiv:2412.01175 [cs]},
	publisher = {arXiv},
	shorttitle = {{OBI}-{Bench}},
	title = {{OBI}-{Bench}: {Can} {LMMs} {Aid} in {Study} of {Ancient} {Script} on {Oracle} {Bones}?},
	url = {http://arxiv.org/abs/2412.01175},
	urldate = {2025-08-09},
	year = {2025}
}

@misc{chen_pictobi-20k_2025,
	author = {Chen, Zijian and Hua, Wenjie and Li, Jinhao and Deng, Lirong and Du, Fan and Chen, Tingzhu and Zhai, Guangtao},
	doi = {10.48550/arXiv.2509.05773},
	month = {September},
	note = {arXiv:2509.05773 [cs]},
	publisher = {arXiv},
	shorttitle = {{PictOBI}-20k},
	title = {{PictOBI}-20k: {Unveiling} {Large} {Multimodal} {Models} in {Visual} {Decipherment} for {Pictographic} {Oracle} {Bone} {Characters}},
	url = {http://arxiv.org/abs/2509.05773},
	urldate = {2025-10-02},
	year = {2025}
}

@article{chen_qin_2025,
	abstract_zh = {Transformer是目前文字识别的主流模型,然而,模型在秦简文字识别领域研究较少。基于自注意力机制的Transformer模型在文字识别中虽然具有较好的性能,但它耗费的计算资源过大,训练过程中容易导致模型过拟合。特别是当文字样本较少时,训练效果较差且更耗费时间。为解决秦简文字识别研究过程中计算量大、准确率低、样本较少等问题,首先构建了国内首个秦简文字单字样本库(共计66973张),并提出了一种基于轻量级的秦简文字识别模型QBSC Transformer(Qin Bamboo Slips Character Recognition Based on Lightweight Transformer)的识别算法。模型采用可分离卷积与窗口自注意力机制融合方式,提升了自注意力机制的泛化能力,可分离卷积提取秦简文字局部特征,窗口自注意力机制提取秦简文字全局特征。基于实际数据的测试结果表明,在秦简文字样本库的识别任务中,QBSC Transformer比目前主流Swin Transformer的准确率提高了0.45\%(达99.46\%),其参数量只有12.7M,计算量为3.4GFlops。},
	author = {Chen, Ming and Chen, Bingquan and Xia, Rong},
	author_zh = {{陈明} and {陈炳权} and {夏蓉}},
	issn = {1006-9348},
	journal = {Computer Simulation},
	journal_zh = {计算机仿真},
	keywords_zh = {深度学习, 秦简文字, 自注意力机制},
	language = {zh-CN},
	number = {4},
	pages = {459--467},
	title = {Qin Bamboo Slips Character Recognition Algorithm Based On Lightweight Transformer Model},
	title_zh = {{基于轻量级Transformer模型的秦简文字识别算法}},
	url = {https://link.cnki.net/urlid/11.3724.TP.20231220.1548.002},
	urldate = {2025-10-02},
	volume = {42},
	year = {2025}
}

@article{chen_recognition_2025,
	abstract_zh = {楚系简帛文字的释读一直是古文字学的重点研究方向，然而目前多依赖人工手段对单字形体开展分析，缺少用计算机视觉技术对海量文字图版进行字形识别的尝试。研究针对大量楚系简帛文字图像识别困难的问题，结合楚系简帛文字的内在特征，不局限于单一深度神经网络模型和单一文字图片分析的微观视角，提出了一种基于集成学习策略的楚系简帛文字图像分类方法，即使用四种深度学习网络提取楚系简帛文字图像的共同形态学特征，并以投票形式得到最终的分类结果，从而构建了计算机自动高效识别海量楚系简帛文字图像的技术框架。应用该框架对目前出土的部分简帛材料中的文字图像进行识别，准确率高达96.72\%,充分证明了该框架的可行性和有效性，为古文字研究提供了新的路径。},
	author = {Chen, Chao and Li, Hezi and Yang, Zekun},
	author_zh = {陈超 and 李赫孜 and 杨泽坤},
	doi = {CNKI:SUN:SZYH.0.2025-02-004},
	issn = {2096-9155},
	journal = {Digital Humanities Research},
	journal_zh = {数字人文研究},
	keywords_zh = {深度学习, 卷积神经网络, 古文字识别, 楚系简帛文字, 集成学习},
	language = {zh-CN},
	number = {2},
	pages = {45--58},
	title = {Recognition of Chu Dynasty characters in Warring States based on Deep Ensemble Learning},
	title_zh = {基于深度集成学习的战国楚系简帛文字识别},
	url = {https://link.cnki.net/doi/CNKI:SUN:SZYH.0.2025-02-004},
	urldate = {2025-10-02},
	volume = {5},
	year = {2025}
}

@inproceedings{diao2023accid,
	author = {Xiaolei Diao and
Daqian Shi and
Jian Li and
others},
	booktitle = {ACMMM},
	pages = {6869--6877},
	title = {Toward Zero-shot Character Recognition: {A} Gold Standard Dataset
with Radical-level Annotations},
	year = {2023}
}

@inproceedings{diao2023rzcr,
	author = {Xiaolei Diao and
Daqian Shi and
Hao Tang and
others},
	booktitle = {IJCAI},
	pages = {654--662},
	title = {{RZCR:} Zero-shot Character Recognition via Radical-based Reasoning},
	year = {2023}
}

@misc{diao_ancient_2025,
	author = {Diao, Xiaolei and Bo, Rite and Xiao, Yanling and Shi, Lida and Zhou, Zhihan and Xu, Hao and Li, Chuntao and Tang, Xiongfeng and Poesio, Massimo and John, Cédric M. and Shi, Daqian},
	doi = {10.48550/arXiv.2506.19208},
	month = {June},
	note = {arXiv:2506.19208 [cs]},
	publisher = {arXiv},
	shorttitle = {Ancient {Script} {Image} {Recognition} and {Processing}},
	title = {Ancient {Script} {Image} {Recognition} and {Processing}: {A} {Review}},
	url = {http://arxiv.org/abs/2506.19208},
	urldate = {2025-06-26},
	year = {2025}
}

@misc{diao_intechar_2025,
	author = {Diao, Xiaolei and Zhou, Zhihan and Shi, Lida and Wang, Ting and Qi, Ruihua and Xu, Hao and Shi, Daqian},
	doi = {10.48550/arXiv.2508.15791},
	month = {August},
	note = {arXiv:2508.15791 [cs]},
	publisher = {arXiv},
	shorttitle = {{InteChar}},
	title = {{InteChar}: {A} {Unified} {Oracle} {Bone} {Character} {List} for {Ancient} {Chinese} {Language} {Modeling}},
	url = {http://arxiv.org/abs/2508.15791},
	urldate = {2025-09-29},
	year = {2025}
}

@misc{duan_restoring_2024,
	author = {Duan, Siyu and Wang, Jun and Su, Qi},
	doi = {10.48550/arXiv.2403.06682},
	month = {March},
	note = {arXiv:2403.06682 [cs]},
	publisher = {arXiv},
	shorttitle = {Restoring {Ancient} {Ideograph}},
	title = {Restoring {Ancient} {Ideograph}: {A} {Multimodal} {Multitask} {Neural} {Network} {Approach}},
	url = {http://arxiv.org/abs/2403.06682},
	urldate = {2025-09-27},
	year = {2024}
}

@book{fudan_excavated_2024,
	address = {Shanghai},
	editor = {{Research Center for Excavated Documents and Paleography, Fudan University}},
	editor_zh = {{复旦大学出土文献与古文字研究中心}},
	month = {January},
	publisher = {Zhongxi Book Company},
	publisher_zh = {中西书局},
	title = {A Course on Excavated Documents and Paleography},
	title_zh = {出土文献与古文字教程},
	year = {2024}
}

@article{gnanadesikan2017towards,
	author = {Gnanadesikan, Amalia E},
	journal = {Writing Systems Research},
	number = {1},
	pages = {14--35},
	title = {Towards a typology of phonemic scripts},
	volume = {9},
	year = {2017}
}

@misc{guan_deciphering_2025,
	author = {Guan, Haisu and Yang, Huanxin and Wang, Xinyu and Han, Shengwei and Liu, Yongge and Jin, Lianwen and Bai, Xiang and Liu, Yuliang},
	doi = {10.48550/arXiv.2406.00684},
	month = {March},
	note = {arXiv:2406.00684 [cs]},
	publisher = {arXiv},
	title = {Deciphering {Oracle} {Bone} {Language} with {Diffusion} {Models}},
	url = {http://arxiv.org/abs/2406.00684},
	urldate = {2025-08-09},
	year = {2025}
}

@misc{guan_open_2024,
	author = {Guan, Haisu and Wan, Jinpeng and Liu, Yuliang and Wang, Pengjie and Zhang, Kaile and Kuang, Zhebin and Wang, Xinyu and Bai, Xiang and Jin, Lianwen},
	doi = {10.48550/arXiv.2401.12467},
	month = {February},
	note = {arXiv:2401.12467 [cs]},
	publisher = {arXiv},
	shorttitle = {An open dataset for the evolution of oracle bone characters},
	title = {An open dataset for the evolution of oracle bone characters: {EVOBC}},
	url = {http://arxiv.org/abs/2401.12467},
	urldate = {2025-09-29},
	year = {2024}
}

@article{guo2022improved,
	author = {Guo, Ziyi and Zhou, Zihan and Liu, Bingshuai and Li, Longquan and Jiao, Qingju and Huang, Chenxi and Zhang, Jianwei},
	journal = {Scientific Programming},
	number = {1},
	pages = {7490363},
	publisher = {Wiley Online Library},
	title = {An Improved Neural Network Model Based on Inception-v3 for Oracle Bone Inscription Character Recognition},
	volume = {2022},
	year = {2022}
}

@inproceedings{han2020accv,
	author = {Han, Wei and Ren, Xiutiao and others},
	booktitle = {ACCV},
	pages = {652--668},
	title = {Self-supervised learning of orc-bert augmentor for recognizing few-shot oracle characters},
	year = {2020}
}

@inproceedings{hu_component-level_nodate,
  title={Component-Level Segmentation for Oracle Bone Inscription Decipherment},
  author={Hu, Zhikai and Cheung, Yiu-ming and Zhang, Yonggang and Peiying, Zhang and Ling, Tang Pui},
  booktitle={Proceedings of the AAAI Conference on Artificial Intelligence},
  volume={39},
  number={27},
  pages={28116--28124},
  year={2025}
}

@misc{huang_agtgan_2023,
	author = {Huang, Hongxiang and Yang, Daihui and Dai, Gang and Han, Zhen and Wang, Yuyi and Lam, Kin-Man and Yang, Fan and Huang, Shuangping and Liu, Yongge and He, Mengchao},
	doi = {10.48550/arXiv.2303.07012},
	month = {March},
	note = {arXiv:2303.07012 [cs]},
	publisher = {arXiv},
	shorttitle = {{AGTGAN}},
	title = {{AGTGAN}: {Unpaired} {Image} {Translation} for {Photographic} {Ancient} {Character} {Generation}},
	url = {http://arxiv.org/abs/2303.07012},
	urldate = {2025-10-02},
	year = {2023}
}

@article{huang_construction_2025,
	abstract_zh = {古文字学是以古文字（即"古汉字"）和各种古文字资料为主要研究对象而建立的一门新兴学科。2019年，在甲骨文发现和研究120周年之际，古文字学界开展了系列纪念活动；2020年，国家有关部委启动实施"古文字与中华文明传承发展工程"。古文字学这一冷门"绝学"迎来前所未有的发展机遇。在新时代背景下，回顾古文字学的发展历史，分析当前我国古文字学研究现状、面临的主要问题，并展望未来，对进一步推进古文字学作为一门新兴交叉学科的建构很有必要。},
	author = {Huang, Dekuan},
	author_zh = {黄德宽},
	issn = {2096-1979},
	journal = {Social Sciences Digest, (6)},
	journal_zh = {社会科学文摘},
	keywords_zh = {古文字学, 学科建构, 交叉学科, 新时代},
	language = {zh-CN},
	number = {6},
	pages = {35--37},
	title = {On the Construction of the Discipline of Chinese Paleography},
	title_zh = {新时代古文字学学科建构},
	url = {https://kns.cnki.net/KCMS/detail/detail.aspx?dbcode=CJFQ&dbname=CJFDLAST2025&filename=SHGC202506011},
	urldate = {2025-11-03},
	year = {2025}
}

@inproceedings{icdar2019,
	author = {Zhang, Yi-Kang and Zhang, Heng and others},
	booktitle = {ICDAR},
	pages = {309--314},
	title = {Oracle Character Recognition by Nearest Neighbor Classification with Deep Metric Learning},
	year = {2019}
}

@misc{jiang_oraclesage_2024,
	author = {Jiang, Hanqi and Pan, Yi and Chen, Junhao and Liu, Zhengliang and Zhou, Yifan and Shu, Peng and Li, Yiwei and Zhao, Huaqin and Mihm, Stephen and Howe, Lewis C. and Liu, Tianming},
	doi = {10.48550/arXiv.2411.17837},
	month = {November},
	note = {arXiv:2411.17837 [cs]},
	publisher = {arXiv},
	shorttitle = {{OracleSage}},
	title = {{OracleSage}: {Towards} {Unified} {Visual}-{Linguistic} {Understanding} of {Oracle} {Bone} {Scripts} through {Cross}-{Modal} {Knowledge} {Fusion}},
	url = {http://arxiv.org/abs/2411.17837},
	urldate = {2025-08-09},
	year = {2024}
}

@inproceedings{li2020hwobc,
	author = {Li, Bin and Dai, Qiong and others},
	booktitle = {Journal of Physics: Conference Series},
	number = {1},
	pages = {012050},
	title = {Hwobc-a handwriting oracle bone character recognition database},
	year = {2020}
}

@article{li_aiguided_2024,
	abstract_zh = {本研究运用"知微缀",以人工智能引导人类直觉的交互方式，带动研究者高效地发现新的甲骨缀合。本研究还融入新的可视分析方法，通过提高人的注意力表现进行加速。},
	author = {Li, Shuangjie},
	author_zh = {李霜洁},
	doi = {CNKI:SUN:CUWX.0.2024-02-003},
	issn = {2096-7365},
	journal = {Excavated Documents, (2)},
	journal_zh = {出土文献},
	keywords_zh = {人工智能, "知微缀", 拼缀, 甲骨},
	language = {zh-CN},
	pages = {17--23, 168--169},
	title = {AI-Guided Human Intuition Discovers New Oracle Bone Fragment Rejoinings: Series 21--30},
	title_zh = {人工智能引导人类直觉产生的甲骨新缀第21—30组},
	url = {https://link.cnki.net/doi/CNKI:SUN:CUWX.0.2024-02-003},
	urldate = {2025-10-02},
	year = {2024}
}

@article{li_ancient_2023,
	abstract_zh = {人工智能与古文字学交叉研究十分重要，开展这项研究既需要人工收集和标注大量数据，同时也需结合恰当的技术。在数据处理方面，数据集建设过程中尽量丰富了单字数量以及字图总量。数据中的字图包括拓本和摹本，其中拓本多带有斑点噪声，降低噪声有助于提高文字识别的准确率。数据中古文字隶定体的显示也是要重点解决的问题。在文字自动识别方面，利用了深度学习算法开展智能识别，从实验结果看，准确率达到八成以上，这是在大规模识别任务下达到的效果，证明了利用人工智能技术识别古文字形体是可行的。分析错误数据可以发现，数据量与形近字是影响识别准确率的关键因素。除了识别以外，知识图谱技术也很重要，建设古文字知识图谱一方面可以实现对古文字知识体系的多角度展示；另一方面也可计算字形中偏旁及构形的相似度，智能寻找出字形之间的联系。},
	author = {Li, Chuntao and Zhang, Qian and Xu, Hao and Gao, Jiaying},
	author_zh = {李春桃 and 张骞 and 徐昊 and 高嘉英},
	doi = {10.15939/j.jujsse.2023.02.wx2},
	issn = {0257-2834},
	journal = {Jilin University Journal of Humanities and Social Sciences},
	journal_zh = {吉林大学社会科学学报},
	keywords_zh = {人工智能, 深度学习, 知识图谱, 古文字研究},
	language = {zh-CN},
	number = {2},
	pages = {164--173, 238--239},
	title = {Ancient Chinese Characters Research Based on Artificial Intelligence Technology},
	title_zh = {基于人工智能技术的古文字研究},
	url = {https://link.cnki.net/doi/10.15939/j.jujsse.2023.02.wx2},
	urldate = {2025-10-02},
	volume = {63},
	year = {2023}
}

@misc{li_comprehensive_2024,
	author = {Li, Jing and Chi, Xueke and Wang, Qiufeng and Wang, Dahan and Huang, Kaizhu and Liu, Yongge and Liu, Cheng-lin},
	doi = {10.48550/arXiv.2411.11354},
	month = {November},
	note = {arXiv:2411.11354 [cs]},
	publisher = {arXiv},
	shorttitle = {A comprehensive survey of oracle character recognition},
	title = {A comprehensive survey of oracle character recognition: challenges, benchmarks, and beyond},
	url = {http://arxiv.org/abs/2411.11354},
	urldate = {2025-08-09},
	year = {2024}
}

@article{li_deep_2018,
	abstract_zh = {考古出土的青铜器铭文是非常宝贵的文字材料,准确、快速地了解其释义和字形演变源流对考古学、历史学和语言学研究均有重要意义.青铜器铭文的辨识需要综合文字的形、音、义进行研究,其中第一步也是最重要的一步就是分析文字的形体特征.本文提出一种基于两阶段特征映射的神经网络模型来提取每个文字的形体特征,最后对比目前已知的文字研究成果,如《古文字类编》、《说文解字》,得出识别的结果.通过定性和定量的实验分析,我们发现本文提出的方法可达到较高的识别精度.特别地,在前10个预测类别中(Top-10)准确率达到了94.2\%，大幅缩小了考古研究者的搜索推测空间,提高了青铜铭文识别的效率和准确性.},
	author = {Li, Wenying and Cao, Bin and Cao, Chunshui and Huang, Yongzhen},
	author_zh = {李文英 and 曹斌 and 曹春水 and 黄永祯},
	doi = {10.16383/j.aas.2018.c180152},
	issn = {0254-4156},
	journal = {Acta Automatica Sinica},
	journal_zh = {自动化学报},
	keywords_zh = {深度学习, 文字识别, 模式识别, 深度卷积神经网络, 青铜器铭文},
	language = {zh-CN},
	number = {11},
	pages = {2023--2030},
	title = {A Deep Learning Based Method for Bronze Inscription Recognition},
	title_zh = {一种基于深度学习的青铜器铭文识别方法},
	url = {https://link.cnki.net/doi/10.16383/j.aas.2018.c180152},
	urldate = {2025-10-02},
	volume = {44},
	year = {2018}
}

@misc{li_diff-oracle_2024,
	author = {Li, Jing and Wang, Qiu-Feng and Wang, Siyuan and Zhang, Rui and Huang, Kaizhu and Cambria, Erik},
	doi = {10.48550/arXiv.2312.13631},
	month = {July},
	note = {arXiv:2312.13631 [cs]},
	publisher = {arXiv},
	shorttitle = {Diff-{Oracle}},
	title = {Diff-{Oracle}: {Deciphering} {Oracle} {Bone} {Scripts} with {Controllable} {Diffusion} {Model}},
	url = {http://arxiv.org/abs/2312.13631},
	urldate = {2025-10-02},
	year = {2024}
}

@article{li_diviner_2021,
	abstract_zh = {董作宾先生的贞人系联法在建构殷墟甲骨文年代框架的过程中,发挥了极为重要的作用,但其中也存在一些不清楚、不准确的问题,致使其年代框架亦有瑕疵,这是学界长期存在争议的焦点。本文旨在接续前人工作,在当前条件下,尽量全面、细密地重新清理甲骨卜辞中的贞人共版关系,重新审视这一问题。本文借助数据建模及可视化分析等手段,是结合数据科学进行出土文献整理与研究的一次尝试。经过重新整理,甲骨卜辞中最大的三个贞人组宾、出、何彻底系联在了一起;本文还从现存材料背后挖掘出多种新的贞人共时关系。在人物共时这一问题上,用完全图模型交叉确定多人共时关系,很大程度上可弥补通行的系联法不准确的缺憾。},
	author = {Li, Shuangjie},
	author_zh = {李霜洁},
	doi = {CNKI:SUN:CUWX.0.2021-04-005},
	issn = {2096-7365},
	journal = {Excavated Documents, (4)},
	journal_zh = {出土文献},
	keywords_zh = {可视化, 数据分析, 甲骨卜辞, 社会网络, 貞人系联},
	language = {zh-CN},
	pages = {44--58, 155},
	title = {Diviner Networks in the Yinxu Oracle Bone Inscriptions: with a Discussion on the Use of the Complete Graph Model in Graph Theory to Cross-determine Multi-person Synchronic Relationships},
	title_zh = {殷墟卜辞中的贞人网络——兼论运用图论中的完全图来交叉确定多人共时关系},
	url = {https://link.cnki.net/doi/CNKI:SUN:CUWX.0.2021-04-005},
	urldate = {2025-10-02},
	year = {2021}
}

@misc{li_mitigating_2025,
	author = {Li, Jinhao and Chen, Zijian and Jiang, Runze and Chen, Tingzhu and Wang, Changbo and Zhai, Guangtao},
	doi = {10.48550/arXiv.2504.09555},
	month = {April},
	note = {arXiv:2504.09555 [cs]},
	publisher = {arXiv},
	shorttitle = {Mitigating {Long}-tail {Distribution} in {Oracle} {Bone} {Inscriptions}},
	title = {Mitigating {Long}-tail {Distribution} in {Oracle} {Bone} {Inscriptions}: {Dataset}, {Model}, and {Benchmark}},
	url = {http://arxiv.org/abs/2504.09555},
	urldate = {2025-08-09},
	year = {2025}
}

@article{li_new_2024,
	abstract_zh = {如今,出土古文字文獻的整理與研究日趨"精密化"和"立體化"。如何持續作有用的"精密化"和"立體化",是一個值得不斷探索和努力的方向。本研究基於李霜潔研發的"支點(LeverX)"古文字文獻數智整理系統,以多種數字化與智能化的計算工具,融合文本-圖像(語言-視覺)多模態信息,構建古文字文獻整理的全新流程與形態,在重新整理殷墟花園莊東地甲骨刻辭的研究實踐上,實現品質與規模的雙重提升。在整理成果方面,蔣玉斌寫定殷墟花園莊東地甲骨刻辭的最新釋文,王子楊摹寫殷墟花園莊東地甲骨的最新字形,在此基础上,本研究進而推出甲骨命名實體標注(Jiagu Named Entity Annotation)、貞卜焦點標注(Divinatory Focus Annotation)、刻辭部位可視化(Inscription Layout Visualization)、辭兆關係圖(Inscription-Crack Graph)、辭際關係圖(Inscription Mapping Graph)等多種面向甲骨文重點疑难問題的全新體例,由此產生殷墟花園莊東地甲骨刻辭新的整理本、文字編和類纂。經過實踐驗證,這套新的流程和工作模式在整理出土古文字文獻材料時,除了在"精密化"和"立體化"方面表現突出,以順應領域研究的發展需求外,還大幅提高原有的速度和減少重複勞動,有利於研究者將工作重心專注於深化内容、提升品質與拓展學術洞察,故具有比較大的優勢,值得進一步挖掘其價值。},
	author = {Li, Shuangjie and Jiang, Yubin and Wang, Ziyang and Liu, Zhiyuan and Sun, Maosong},
	author_zh = {李霜潔 and 蔣玉斌 and 王子楊 and 劉知遠 and 孫茂松},
	doi = {CNKI:SUN:GUDX.0.2024-01-007},
	journal = {The Journal of Chinese Classical Studies},
	journal_zh = {中国古典学},
	keywords_zh = {出土古文字文獻整理, 刻辭部位, 甲骨命名實體, 計算甲骨學, 貞卜焦點, 辭兆關係圖, 辭際關係圖},
	language = {zh-CN},
	number = {1},
	pages = {67--86},
	title = {A New Digital-Intelligence-Enhanced Organization of Ancient Script Documents: A Case Study of the Oracle Bone Inscriptions from Huayuanzhuang Dongdi, Yinxu},
	title_zh = {數智增强的古文字文獻新整理:以殷墟花園莊東地甲骨刻辭爲例},
	url = {https://link.cnki.net/doi/CNKI:SUN:GUDX.0.2024-01-007},
	urldate = {2025-10-02},
	volume = {5},
	year = {2024}
}

@misc{li_obiformer_2025,
	author = {Li, Jinhao and Chen, Zijian and Chen, Tingzhu and Liu, Zhiji and Wang, Changbo},
	doi = {10.48550/arXiv.2504.13524},
	month = {April},
	note = {arXiv:2504.13524 [cs]},
	publisher = {arXiv},
	shorttitle = {{OBIFormer}},
	title = {{OBIFormer}: {A} {Fast} {Attentive} {Denoising} {Framework} for {Oracle} {Bone} {Inscriptions}},
	url = {http://arxiv.org/abs/2504.13524},
	urldate = {2025-08-09},
	year = {2025}
}

@misc{li_oracle_2024,
	author = {Li, Bang and Luo, Donghao and Liang, Yujie and Yang, Jing and Ding, Zengmao and Peng, Xu and Jiang, Boyuan and Han, Shengwei and Sui, Dan and Qin, Peichao and Wu, Pian and Wang, Chaoyang and Qi, Yun and Jin, Taisong and Wang, Chengjie and Huang, Xiaoming and Shu, Zhan and Ji, Rongrong and Liu, Yongge and Wu, Yunsheng},
	doi = {10.48550/arXiv.2407.03900},
	month = {July},
	note = {arXiv:2407.03900 [cs]},
	publisher = {arXiv},
	title = {Oracle {Bone} {Inscriptions} {Multi}-modal {Dataset}},
	url = {http://arxiv.org/abs/2407.03900},
	urldate = {2025-08-09},
	year = {2024}
}

@misc{li_oraclefusion_2025,
	author = {Li, Caoshuo and Ding, Zengmao and Hu, Xiaobin and Li, Bang and Luo, Donghao and Wu, AndyPian and Wang, Chaoyang and Wang, Chengjie and Jin, Taisong and SevenShu and Wu, Yunsheng and Liu, Yongge and Ji, Rongrong},
	doi = {10.48550/arXiv.2506.21101},
	month = {June},
	note = {arXiv:2506.21101 [cs]},
	publisher = {arXiv},
	shorttitle = {{OracleFusion}},
	title = {{OracleFusion}: {Assisting} the {Decipherment} of {Oracle} {Bone} {Script} with {Structurally} {Constrained} {Semantic} {Typography}},
	url = {http://arxiv.org/abs/2506.21101},
	urldate = {2025-08-09},
	year = {2025}
}

@article{lin2022radical,
	author = {Lin, Xiaoyu and Chen, Siyuan and Zhao, Feng and Qiu, Xipeng},
	journal = {International Journal on Document Analysis and Recognition},
	number = {3},
	pages = {219--235},
	title = {Radical-based extract and recognition networks for oracle character recognition},
	volume = {25},
	year = {2022}
}

@incollection{liu1995script,
	author = {Liu, In-Mao},
	booktitle = {Scripts and literacy: Reading and learning to read alphabets, syllabaries and characters},
	pages = {145--162},
	publisher = {Springer},
	title = {Script factors that affect literacy: Alphabetic vs. logographic languages},
	year = {1995}
}

@inproceedings{liu_recognition_2021,
	author = {Liu, Guoying and Ge, Wenying and Du, Bingxin},
	booktitle = {2021 {IEEE} 4th {International} {Conference} on {Information} {Systems} and {Computer} {Aided} {Education} ({ICISCAE})},
	doi = {10.1109/ICISCAE52414.2021.9590692},
	month = {September},
	pages = {39--42},
	title = {Recognition of {OBIC}'s {Variants} by {Using} {Deep} {Neural} {Networks} and {Spectral} {Clustering}},
	url = {https://ieeexplore.ieee.org/abstract/document/9590692},
	urldate = {2025-10-02},
	year = {2021}
}

@article{luxuzheng2020,
	author = {Lu, Xuzheng and Cai, Hong and Lin, Li},
	journal = {CAAI Transactions on Intelligent Systems},
	number = {2},
	pages = {12},
	title = {Recognition of oracle radical based on the capsule network (in chinese)},
	volume = {15},
	year = {2020}
}

@article{mai_oracle_2024,
	author = {Mai, Christopher and Penava, Pascal and Buettner, Ricardo},
	doi = {10.1109/ACCESS.2024.3521319},
	issn = {2169-3536},
	journal = {IEEE Access},
	pages = {197021--197034},
	title = {Oracle {Bone} {Inscription} {Character} {Recognition} {Based} on a {Novel} {Convolutional} {Neural} {Network} {Architecture}},
	url = {https://ieeexplore.ieee.org/document/10811903/},
	urldate = {2025-09-29},
	volume = {12},
	year = {2024}
}

@inproceedings{meng2017recognition,
	author = {Meng, Lin},
	booktitle = {ICPRAM},
	pages = {606--611},
	title = {Recognition of Oracle Bone Inscriptions by Extracting Line Features on Image Processing.},
	year = {2017}
}

@article{mo2022computational,
	author = {Mo, Bofeng},
	author_zh = {莫伯峰},
	day = {30},
	journal = {Guangming Daily},
	journal_zh = {光明日报},
	month = {October},
	title = {'Computational Paleography' Is on Its Way},
	title_zh = {"计算古文字学"正在路上},
	url = {https://news.gmw.cn/2022-10/30/content_36123647.htm},
	year = {2022}
}

@inproceedings{OBC306,
	author = {Huang, Shuangping and Wang, Haobin and Liu, Yongge and others},
	booktitle = {ICDAR},
	pages = {681--688},
	title = {{OBC306:} {A} Large-Scale Oracle Bone Character Recognition Dataset},
	year = {2019}
}

@article{ou2024qin,
	author = {Ou, Yun and Zhou, Zhen-Jie and Kang, Di-Wen and Zhou, Pan and Liu, Xue-Wei},
	journal = {Baghdad Science Journal},
	number = {2 (SI)},
	pages = {0696--0696},
	title = {Qin Seal Script Character Recognition with Fuzzy and Incomplete Information},
	volume = {21},
	year = {2024}
}

@article{peng_interpretable_2025,
	author = {Peng, Kaixin and Zhao, Mengyang and Yu, Haiyang and Fu, Teng and Li, Bin},
	doi = {10.48550/arXiv.2508.10113},
	month = {August},
	note = {arXiv:2508.10113 [cs]},
	publisher = {arXiv},
	title = {Interpretable {Oracle} {Bone} {Script} {Decipherment} through {Radical} and {Pictographic} {Analysis} with {LVLMs}},
	url = {http://arxiv.org/abs/2508.10113},
	urldate = {2025-10-02},
	year = {2025}
}

@article{qi_ancientglyphnet_2025,
	author = {Qi, Hengnian and Yang, Hao and Wang, Zhaojiang and Ye, Jiabin and Xin, Qiuyi and Zhang, Chu and Lang, Qing},
	doi = {10.1007/s10462-024-11095-5},
	issn = {1573-7462},
	journal = {Artificial Intelligence Review},
	language = {en},
	month = {January},
	number = {3},
	pages = {88},
	shorttitle = {{AncientGlyphNet}},
	title = {{AncientGlyphNet}: an advanced deep learning framework for detecting ancient {Chinese} characters in complex scene},
	url = {https://link.springer.com/10.1007/s10462-024-11095-5},
	urldate = {2025-09-29},
	volume = {58},
	year = {2025}
}

@inproceedings{qiao_v-oracle_2025,
	address = {Vienna, Austria},
	author = {Qiao, Runqi and Tan, Qiuna and Dong, Guanting and MinhuiWu, MinhuiWu and Wang, Jiapeng and Zhang, YiFan and GongQue, Zhuoma and Sun, Chong and Xu, Yida and Xue, Yadong and Tian, Ye and Bao, Zhimin and Yang, Lan and Li, Chen and Zhang, Honggang},
	booktitle = {Proceedings of the 63rd {Annual} {Meeting} of the {Association} for {Computational} {Linguistics} ({Volume} 1: {Long} {Papers})},
	doi = {10.18653/v1/2025.acl-long.986},
	editor = {Che, Wanxiang and Nabende, Joyce and Shutova, Ekaterina and Pilehvar, Mohammad Taher},
	isbn = {979-8-89176-251-0},
	month = {July},
	pages = {20124--20150},
	publisher = {Association for Computational Linguistics},
	shorttitle = {V-{Oracle}},
	title = {V-{Oracle}: {Making} {Progressive} {Reasoning} in {Deciphering} {Oracle} {Bones} for {You} and {Me}},
	url = {https://aclanthology.org/2025.acl-long.986/},
	urldate = {2025-09-29},
	year = {2025}
}

@book{qiu2000chinese,
	address = {New Haven, CT},
	author = {Chiu, Hsi-kuei},
	isbn = {1-55729-071-7},
	note = {Translated by Gilbert L. Mattos and Jerry Norman. Translation of \textit{Wenzixue gaiyao}},
	number = {4},
	publisher = {The Society for the Study of Early China},
	series = {Early China Special Monograph Series},
	title = {Chinese Writing},
	year = {2000}
}

@inproceedings{roy2009seal,
	author = {Roy, Partha Pratim and Pal, Umapada and Llad{\'o}s, Josep},
	booktitle = {2009 10th International Conference on Document Analysis and Recognition},
	organization = {IEEE},
	pages = {101--105},
	title = {Seal detection and recognition: An approach for document indexing},
	year = {2009}
}

@inproceedings{shi2022rcrn,
	author = {Daqian Shi and
Xiaolei Diao and
others},
	booktitle = {ACMMM},
	pages = {1177--1185},
	title = {{RCRN:} Real-world Character Image Restoration Network via Skeleton
Extraction},
	year = {2022}
}

@misc{shi_cola_2025,
	author = {Shi, Fan and Yu, Haiyang and Li, Bin and Xue, Xiangyang},
	doi = {10.48550/arXiv.2506.03798},
	month = {June},
	note = {arXiv:2506.03798 [cs]},
	publisher = {arXiv},
	shorttitle = {{CoLa}},
	title = {{CoLa}: {Chinese} {Character} {Decomposition} with {Compositional} {Latent} {Components}},
	url = {http://arxiv.org/abs/2506.03798},
	urldate = {2025-08-09},
	year = {2025}
}

@inproceedings{sun2020dual,
	author = {Sun, Wenjie and Zhai, Guangtao and Gao, Zhongpai and Chen, Tingzhu and Zhu, Yucheng and Wang, Zhaodi},
	booktitle = {2020 IEEE Conference on Multimedia Information Processing and Retrieval (MIPR)},
	organization = {IEEE},
	pages = {193--198},
	title = {Dual-view oracle bone script recognition system via temporal-spatial psychovisual modulation},
	year = {2020}
}

@misc{tao_clustering-based_2025,
	author = {Tao, Ye and Fu, Xinran and Pang, Honglin and Yang, Xi and Li, Chuntao},
	doi = {10.48550/arXiv.2508.18641},
	month = {August},
	note = {arXiv:2508.18641 [cs]},
	publisher = {arXiv},
	title = {Clustering-based {Feature} {Representation} {Learning} for {Oracle} {Bone} {Inscriptions} {Detection}},
	url = {http://arxiv.org/abs/2508.18641},
	urldate = {2025-09-29},
	year = {2025}
}

@article{wang2024dataset,
	author = {Wang, Mei and Deng, Weihong},
	journal = {Scientific Data},
	number = {1},
	pages = {87},
	title = {A dataset of oracle characters for benchmarking machine learning algorithms},
	volume = {11},
	year = {2024}
}

@article{wang_innovative_2025,
	abstract_zh = {该文提出深度学习技术在甲骨文图像中的应用和改进，旨在提高甲骨文图像的检测与识别的效率和准确性，是一种针对甲骨文图像研究的整体解决方案。甲骨文图像样本量稀少，细节复杂，且分为临摹和拓片两种图像。针对识别样本量稀少的问题，方案创新性提出字体图像类的数据增强WDA算法和拓片化的处理过程；针对拓片甲骨文字检测数据集的空白，方案构建数据集Oracle-Detect；针对传统模型识别率低，难以增量识别的问题，研究改进了基于Stacking集成式的甲骨文字识别模型。实验结果表明，该方案在甲骨文字的检测mPA达到94\%，在临摹甲骨文字识别任务上准确率达到83\%，拓片甲骨文字识别准确率达到81\%。},
	author = {Wang, Zexin and Tong, Hengjian},
	author_zh = {王泽鑫 and 童恒建},
	doi = {10.11907/rjdk.251006},
	issn = {1672-7800},
	journal = {Software Guide},
	journal_zh = {软件导刊},
	keywords_zh = {甲骨文, 图像识别, 数据增强处理, 模型集成, 深度学习, 目标检测, 计算机视觉},
	language = {zh-CN},
	month = {July},
	pages = {1--11},
	title = {Innovative Applications and Improvements of Deep Learning in the Detection and Recognition of Oracle-bone Character Images},
	title_zh = {基于深度学习的甲骨文图像检测与识别创新应用与改进研究},
	url = {https://link.cnki.net/doi/10.11907/rjdk.251006},
	urldate = {2025-10-02},
	year = {2025}
}

@article{wang_open_2024,
	author = {Wang, Pengjie and Zhang, Kaile and Wang, Xinyu and Han, Shengwei and Liu, Yongge and Wan, Jinpeng and Guan, Haisu and Kuang, Zhebin and Jin, Lianwen and Bai, Xiang and Liu, Yuliang},
	doi = {10.1038/s41597-024-03807-x},
	issn = {2052-4463},
	journal = {Scientific Data},
	language = {en},
	month = {September},
	number = {1},
	pages = {976},
	title = {An open dataset for oracle bone character recognition and decipherment},
	url = {https://www.nature.com/articles/s41597-024-03807-x},
	urldate = {2025-09-29},
	volume = {11},
	year = {2024}
}

@misc{wang_puzzle_2024,
	author = {Wang, Pengjie and Zhang, Kaile and Wang, Xinyu and Han, Shengwei and Liu, Yongge and Jin, Lianwen and Bai, Xiang and Liu, Yuliang},
	doi = {10.48550/arXiv.2406.03019},
	month = {June},
	note = {arXiv:2406.03019 [cs]},
	publisher = {arXiv},
	shorttitle = {Puzzle {Pieces} {Picker}},
	title = {Puzzle {Pieces} {Picker}: {Deciphering} {Ancient} {Chinese} {Characters} with {Radical} {Reconstruction}},
	url = {http://arxiv.org/abs/2406.03019},
	urldate = {2025-08-09},
	year = {2024}
}

@article{wang_unsupervised_2022,
	author = {Wang, Mei and Deng, Weihong and Liu, Cheng-Lin},
	doi = {10.1109/TIP.2022.3165989},
	issn = {1057-7149, 1941-0042},
	journal = {IEEE Transactions on Image Processing},
	note = {arXiv:2205.06549 [cs]},
	pages = {3137--3150},
	title = {Unsupervised {Structure}-{Texture} {Separation} {Network} for {Oracle} {Character} {Recognition}},
	url = {http://arxiv.org/abs/2205.06549},
	urldate = {2025-10-02},
	volume = {31},
	year = {2022}
}

@inproceedings{wu2021ancient,
	author = {Wu, Lingjing and Zhang, Chuang and Xu, Mengqiu and Wu, Ming},
	booktitle = {2021 7th IEEE International Conference on Network Intelligence and Digital Content (IC-NIDC)},
	organization = {IEEE},
	pages = {309--313},
	title = {Ancient Chinese recognition method based on attention mechanism},
	year = {2021}
}

@inproceedings{wu_cnn-based_2022,
	author = {Wu, Xuanqi and Wang, Ziyang and Ren, Peng},
	booktitle = {2022 5th {International} {Conference} on {Advanced} {Electronic} {Materials}, {Computers} and {Software} {Engineering} ({AEMCSE})},
	doi = {10.1109/AEMCSE55572.2022.00106},
	month = {April},
	pages = {514--519},
	title = {{CNN}-based {Bronze} {Inscriptions} {Character} {Recognition}},
	url = {https://ieeexplore.ieee.org/abstract/document/9948332},
	urldate = {2025-10-02},
	year = {2022}
}

@misc{wu_cross-font_2024,
	author = {Wu, Zhicong and Su, Qifeng and Gu, Ke and Shi, Xiaodong},
	doi = {10.48550/arXiv.2409.06381},
	month = {December},
	note = {arXiv:2409.06381 [cs]},
	publisher = {arXiv},
	title = {A {Cross}-{Font} {Image} {Retrieval} {Network} for {Recognizing} {Undeciphered} {Oracle} {Bone} {Inscriptions}},
	url = {http://arxiv.org/abs/2409.06381},
	urldate = {2025-08-09},
	year = {2024}
}

@book{xu_shuowen_1963,
	address = {Beijing},
	author = {Xu, Shen},
	author_zh = {许慎},
	publisher = {Zhonghua Shuju},
	publisher_zh = {中华书局},
	title = {Shuowen Jiezi Zhu},
	title_zh = {说文解字注},
	year = {1963}
}

@article{miao_research_2022,
	author = {Miao, Yalin and Liang, Li and Ji, Yichun and Li, Guodong},
	doi = {10.1038/s41598-022-24388-y},
	issn = {2045-2322},
	journal = {Scientific Reports},
	language = {en},
	month = {November},
	number = {1},
	pages = {19795},
	title = {Research on denoising method of chinese ancient character image based on chinese character writing standard model},
	url = {https://www.nature.com/articles/s41598-022-24388-y},
	urldate = {2025-09-29},
	volume = {12},
	year = {2022}
}

@article{yue2022obi125,
	author = {Yue, Xuebin and Li, Hengyi and Fujikawa, Yoshiyuki and Meng, Lin},
	journal = {Journal on Computing and Cultural Heritage},
	number = {4},
	pages = {1--20},
	title = {Dynamic Dataset Augmentation for Deep Learning-based Oracle Bone Inscriptions Recognition},
	volume = {15},
	year = {2022}
}

@inproceedings{zhang2019oracle,
	author = {Zhang, Yi-Kang and Zhang, Heng and Liu, Yong-Ge and Yang, Qing and Liu, Cheng-Lin},
	booktitle = {2019 International Conference on Document Analysis and Recognition (ICDAR)},
	organization = {IEEE},
	pages = {309--314},
	title = {Oracle character recognition by nearest neighbor classification with deep metric learning},
	year = {2019}
}

@inproceedings{zhang2021ancient35,
	author = {Zhang, Guanyu and Liu, Dequan and others},
	booktitle = {ICCBR},
	pages = {309--324},
	title = {Deciphering ancient chinese oracle bone inscriptions using case-based reasoning},
	year = {2021}
}

@misc{zhang_explainable_2025,
	author = {Zhang, Chongsheng and Wu, Shuwen and Chen, Yingqi and Men, Yi and Fan, Gaojuan and Aßenmacher, Matthias and Heumann, Christian and Gama, João},
	doi = {10.48550/arXiv.2505.03836},
	month = {July},
	note = {arXiv:2505.03836 [cs]},
	publisher = {arXiv},
	title = {Explainable {Coarse}-to-{Fine} {Ancient} {Manuscript} {Duplicates} {Discovery}},
	url = {http://arxiv.org/abs/2505.03836},
	urldate = {2025-08-09},
	year = {2025}
}

@article{zhang_oracle_2023,
	abstract_zh = {甲骨残片缀合一直是甲骨学研究中最急迫最具基础性的工作，它使得甲骨残片经过拼接，复原为更加完整的原始材料.尽管前人及同行曾提出若干计算机辅助的甲骨缀合方法，但这些方法缀合准确度不足，未能真正投入使用，并不能真正帮助专家解决甲骨缀合问题，导致当前的甲骨缀合工作仍旧依靠人工、依旧费时费力.为了更好地研究甲骨残片的机器缀合问题，本文使用一个较大规模甲骨缀合基准数据集OB-Rejoin，该数据集包含了约一千幅甲骨拓片图像，且融入了大量的甲骨学界已缀成果，用于算法评估.基于该数据集，本文设计了一种基于斜率变化量序列匹配的甲骨缀合算法（Slope United Sequence Matching for Oracle Bone Fragments Conjugation,SUM），该方法将甲骨残片的断边碴口图像匹配问题转化为数值型的序列数据和序列相似性比对问题，以将尚不够非常精密的计算机视觉领域的碴口图像匹配问题转换为数据科学领域较为成熟的序列数据相似性匹配问题. SUM将数值型的碴口序列数据进一步转换为斜率变化量序列和字符序列数据，最后利用字符序列的模糊匹配完成甲骨残片的碴口匹配.在实验环节，SUM算法与经典的序列相似性计算方法在精确率、召回率、漏检率方面进行了对比，并与两个较新的基于深度学习的序列匹配算法和形状匹配算法进行了性能对比.整体而言，SUM在OB-Rejoin数据集上的Top-15缀合召回率达到了95.181\%，超越了对比算法.重要出土文献的精准复原本身是历史学和古文字研究中客观存在的重大现实需求，具有重要的史学价值和意义，因此，本文的研究成果，不但有助于解决甲骨残片的机器缀合问题，还对秦汉简牍和敦煌遗书等重要出土文献的精准复原具有重要的参考价值.},
	author = {Zhang, Chongsheng and Wang, Bin},
	author_zh = {张重生 and 王斌},
	issn = {0372-2112},
	journal = {Acta Electronica Sinica},
	journal_zh = {电子学报},
	keywords_zh = {序列相似性计算, 形状匹配, 甲骨文, 甲骨缀合, 边缘匹配},
	language = {zh-CN},
	number = {4},
	pages = {860--869},
	title = {Oracle Bone Fragments Conjugation Based on Sequence Matching},
	title_zh = {基于序列相似性计算的甲骨残片缀合算法},
	url = {https://link.cnki.net/urlid/11.2087.tn.20230303.1121.086},
	urldate = {2025-10-02},
	volume = {51},
	year = {2023}
}

@article{zhou2023style,
	author = {Zhou, Wenhui and Liu, Jinyu and Li, Jiefeng and Li, Jiyi and Lin, Lili and Fukumoto, Fumiyo and Dai, Guojun},
	journal = {Journal of the Franklin Institute},
	number = {16},
	pages = {11295--11313},
	publisher = {Elsevier},
	title = {Style-independent radical sequence learning for zero-shot recognition of Small Seal script},
	volume = {360},
	year = {2023}
}

@misc{zhu_rejoining_2025,
	author = {Zhu, Jinchi and Zhao, Zhou and Lei, Hailong and Wang, Xiaoguang and Lu, Jialiang and Li, Jing and Tang, Qianqian and Shen, Jiachen and Xia, Gui-Song and Du, Bo and Xu, Yongchao},
	doi = {10.48550/arXiv.2505.08601},
	month = {July},
	note = {arXiv:2505.08601 [cs]},
	publisher = {arXiv},
	title = {Rejoining fragmented ancient bamboo slips with physics-driven deep learning},
	url = {http://arxiv.org/abs/2505.08601},
	urldate = {2025-09-29},
	year = {2025}
}

@article{zhou1995jia,
  author       = {Xin{-}Lun Zhou and
                  Xing{-}Cheng Hua and
                  Feng Li},
  title        = {A method of Jia Gu Wen recognition based on a two-level classification},
  booktitle    = {ICDAR},
  pages        = {833--836},
  year         = {1995}
}

@article{li2011isomorphism,
  title={Recognition of Inscriptions on Bones or Tortoise Shells Based on Graph Isomorphism (in Chinese)},
  author={Qingsheng Li and 
          Xingyu Yang and 
          Aimin Wang},
  journal={Computer Engineering and Applications},
  volume={47},
  number={8},
  pages={3},
  year={2011},
}

@article{luo2023aggregation,
  author       = {Yanlong Luo and
                  Yiwen Sun and
                  Xiaojun Bi},
  title        = {Multiple attentional aggregation network for handwritten Dongba character
                  recognition},
  journal      = {Expert Systems with Applications},
  volume       = {213},
  pages        = {118865},
  year         = {2023}
}

@inproceedings{li2023decouple,
  author       = {Jing Li and
                  Bin Dong and
                  Qiu{-}Feng Wang and
                  others},
  title        = {Decoupled Learning for Long-Tailed Oracle Character Recognition},
  booktitle    = {ICDAR},
  pages        = {165--181},
  year         = {2023}
}

@inproceedings{qiao2024making,
  title={Making Visual Sense of Oracle Bones for You and Me},
  author={Qiao, Runqi and 
          Yang, Lan and 
          Pang, Kaiyue and 
          Zhang, Honggang},
  booktitle={CVPR},
  pages={12656--12665},
  year={2024}
}

@article{shi2017chinese,
  title={A Chinese character structure preserved denoising method for Chinese tablet calligraphy document images based on KSVD dictionary learning},
  author={Shi, Zhenghao and Xu, Binxin and Zheng, Xia and Zhao, Minghua},
  journal={Multimedia Tools and Applications},
  volume={76},
  pages={14921--14936},
  year={2017},
  publisher={Springer}
}

@article{huang2016comparison,
  title={Comparison of different image denoising algorithms for Chinese calligraphy images},
  author={Huang, Zhi-Kai and Li, Zhi-Hong and Huang, Han and Li, Zhi-Biao and Hou, Ling-Ying},
  journal={Neurocomputing},
  volume={188},
  pages={102--112},
  year={2016},
  publisher={Elsevier}
}

@article{li2023towards,
  title={Towards better long-tailed oracle character recognition with adversarial data augmentation},
  author={Li, Jing and Wang, Qiu-Feng and Huang, Kaizhu and Yang, Xi and Zhang, Rui and Goulermas, John Y},
  journal={Pattern Recognition},
  volume={140},
  pages={109534},
  year={2023},
  publisher={Elsevier}
}

@inproceedings{zhang2021bag,
  title={Bag of tricks for long-tailed visual recognition with deep convolutional neural networks},
  author={Zhang, Yongshun and Wei, Xiu-Shen and Zhou, Boyan and Wu, Jianxin},
  booktitle={Proceedings of the AAAI conference on artificial intelligence},
  volume={35},
  number={4},
  pages={3447--3455},
  year={2021}
}

@article{WOS001150626700002,
  author = {Yuan, J and Chen, SX and Mo, BF and Ma, YQ and Zheng, WJ and Zhang, CS},
  title = {R-GNN: recurrent graph neural networks for font classification of oracle bone inscriptions},
  journal = {HERITAGE SCIENCE},
  year = {2024},
  volume = {12},
  number = {1},
  doi = {10.1186/s40494-024-01133-4},
  month = {jan},
  articleno = {30},
}

@article{WOS001204747000001,
  author = {Gao, F and Chen, X and Li, B and Liu, YG and Jiang, RH and Han, YH},
  title = {Linking unknown characters via oracle bone inscriptions retrieval},
  journal = {MULTIMEDIA SYSTEMS},
  year = {2024},
  volume = {30},
  number = {3},
  doi = {10.1007/s00530-024-01327-7},
  month = {jun},
  articleno = {125},
}

@article{WOS000663706800002,
  author = {Xiong, J and Liu, GY and Liu, YG and Liu, MT},
  title = {Oracle Bone Inscriptions information processing based on multi-modal knowledge graph},
  journal = {COMPUTERS AND ELECTRICAL ENGINEERING},
  year = {2021},
  volume = {92},
  doi = {10.1016/j.compeleceng.2021.107173},
  month = {jun},
  articleno = {107173},
}

@article{WOS000564295200001,
  author = {Gao, JH and Liang, X},
  title = {Distinguishing Oracle Variants Based on the Isomorphism and Symmetry Invariances of Oracle-Bone Inscriptions},
  journal = {IEEE ACCESS},
  year = {2020},
  volume = {8},
  doi = {10.1109/ACCESS.2020.3017533},
  pages = {152258--152275},
}

@article{WOS001102506600004,
  author = {Jiang, Y and Chen, SX and Gao, WZ and Peng, ML and Jiang, LH},
  title = {SFF-Siam: A New Oracle Bone Rejoining Method Based on Siamese Network},
  journal = {IEEE COMPUTER GRAPHICS AND APPLICATIONS},
  year = {2023},
  volume = {43},
  number = {6},
  doi = {10.1109/MCG.2023.3284000},
  month = {nov},
  pages = {22--32},
}

@article{WOS001570216400003,
  author = {Wu, R and Wang, SB and Li, XS and Liu, D},
  title = {A text image dual conditional stable diffusion model for oracle bone inscription decipherment},
  journal = {NPJ HERITAGE SCIENCE},
  year = {2025},
  volume = {13},
  number = {1},
  doi = {10.1038/s40494-025-02019-9},
  month = {sep},
  articleno = {453},
}

@article{WOS001514394500002,
  author = {Li, B and Ding, ZM and Zhang, Y and Zhang, H and Yang, J and Fang, JQ and Jin, TS and Liu, YG and Song, ZH and Ji, RR},
  title = {An open benchmark for oracle bone rubbing image retrieval},
  journal = {NPJ HERITAGE SCIENCE},
  year = {2025},
  volume = {13},
  number = {1},
  doi = {10.1038/s40494-025-01859-9},
  month = {jun},
  articleno = {292},
}

@article{zh-mo_application__2023,
	title = {The Application and Prospect of Artificial Intelligence in the Study of Paleography},
	title_zh = {人工智能在古文字研究中的应用及展望},
	issn = {1005-3247},
	url = {https://link.cnki.net/doi/10.15990/j.cnki.cn11-3306/g2.2023.02.012},
	doi = {10.15990/j.cnki.cn11-3306/g2.2023.02.012},
	abstract_zh = {本文对近期人工智能在古文字研究中的应用状况进行了回顾和总结。在古文字字形相关的任务中，古文字识别、古文字检测都已经取得了较大进展，可以为古文字研究提供切实的辅助。古文字分类、古文字考释离实际需求还有较大距离，但是发展思路已经逐渐明晰。在古文字材料整理类任务中，甲骨缀合、甲骨校重、青铜器断代等都已经取得了阶段性成果，推进了古文字研究的进步。展望人工智能与古文字交叉研究的未来发展，多模态模型的应用将会是主要发展方向，生成式人工智能的发散思维将为古文字问题解答提供更多思路，而借助大数据智能的数据挖掘能力则可以发现更多古文字中的规律性现象。},
	language = {zh-CN},
	number = {2},
	urldate = {2025-11-16},
	journal = {Chinese Culture Research},
	journal_zh = {中国文化研究},
	author = {Mo, Bofeng and Zhang, Chongsheng},
	author_zh = {{莫伯峰} and {张重生}},
	year = {2023},
	keywords_zh = {人工智能, 古文字, 展望, 应用},
	pages = {47--56},
}

@article{zh-li_ding__2023,
	title = {Periodization of Bronze {Ding-Vessels} Based on Deep Learning},
	title_zh = {基于深度学习技术的青铜鼎分期断代研究},
	issn = {2096-7365},
	url = {https://kns.cnki.net/KCMS/detail/detail.aspx?dbcode=CJFQ&dbname=CJFDLAST2023&filename=CUWX202303003},
	abstract_zh = {分期断代是青铜器研究的重要基础，但铜器断代工作具有较高的专业门槛，一直依赖少数专家人工完成。人工智能的迅速发展，使青铜器智能断代成为可能。本文以青铜鼎为对象，提出利用人工智能深度学习技术对先秦时期青铜器进行断代的方法，并从数据处理、模型搭建、具体实验、结果分析等多个角度展开研究。实验结果表明，人工智能模型能够准确判断绝大多数青銅鼎的时代。同时，研究成果也已转换成实际应用，模型已经部署于微信小程序。},
	language = {zh-CN},
	number = {3},
	urldate = {2025-11-17},
	journal = {Excavated Documents},
	journal_zh = {出土文献},
	author = {Li, Chuntao and Qi, Ruihua and Yang, Xi and Zhou, Rixin},
	author_zh = {{李春桃} and {戚睿华} and {杨溪} and {周日鑫}},
	year = {2023},
	keywords_zh = {人工智能, 应用, 断代, 青铜器},
	pages = {16--32, 154--155},
}

@article{wu_paleography_2020,
    author = {Wu, Zhenwu},
    author_zh = {{吴振武}},
    title = {Paleography Decipherment and Artificial Intelligence},
    title_zh = {古文字考释与人工智能},
    journal = {Guangming Daily},
    year = {2020},
    month = {11},
    day = {7},
    pages = {12}
}

@article{li_how_2022,
    author = {Li, Chuntao},
    author_zh = {{李春桃}},
    title = {How Artificial Intelligence Assists Paleography Research},
    title_zh = {人工智能如何辅助古文字研究},
    journal = {Guangming Daily},
    year = {2022},
    month = {10},
    day = {30},
    pages = {5}
}

@inproceedings{hu_new_2021,
    author = {Hu, Renfen and Zhang, Jing and Zhang, Yifan and Liu, Meng and Zhang, Jia},
    author_zh = {{胡韧奋} and {张晶} and {张一帆} and {刘猛} and {张家}},
    title = {A New Method of Automatic Sentence Segmentation and Punctuation for Ancient Chinese},
    title_zh = {古汉语自动断句与标点新方法},
    booktitle = {Proceedings of the 20th Chinese National Conference on Computational Linguistics},
    year = {2021},
    pages = {959--969}
}

@misc{tang_chisiec_2024,
	title = {{CHisIEC}: {An} {Information} {Extraction} {Corpus} for {Ancient} {Chinese} {History}},
	shorttitle = {{CHisIEC}},
	url = {http://arxiv.org/abs/2403.15088},
	language = {en},
	urldate = {2024-07-17},
	publisher = {arXiv},
	author = {Tang, Xuemei and Deng, Zekun and Su, Qi and Yang, Hao and Wang, Jun},
	month = apr,
	year = {2024},
	note = {arXiv:2403.15088 [cs]},
}

@article{zh-tang_ancient-chinese-word_2023,
	title = {Ancient Chinese Word Segmentation Based on Graph Convolutional Neural Network},
	title_zh = {基于图卷积神经网络的古汉语分词研究},
	author = {Tang, Xuemei and Su, Qi and Wang, Jun and Yang, Hao},
	author_zh = {{唐雪梅} and {苏祺} and {王军} and {杨浩}},
	journal = {Journal of the China Society for Scientific and Technical Information},
	journal_zh = {情报学报},
	volume = {42},
	number = {6},
	pages = {740--750},
	year = {2023},
	issn = {1000-0135},
	url = {https://link.cnki.net/doi/CNKI:SUN:QBXB.0.2023-06-009},
	urldate = {2025-09-27},
	doi = {CNKI:SUN:QBXB.0.2023-06-009},
	abstract_zh = {古汉语的语法有省略、语序倒置的特点，词法有词类活用、代词名词丰富的特点，这些特点增加了古汉语分词的难度，并带来严重的out-of-vocabulary (OOV)问题。目前，深度学习方法已被广泛地应用在古汉语分词任务中并取得了成功，但是这些研究更关注的是如何提高分词效果，忽视了分词任务中的一大挑战，即OOV问题。因此，本文提出了一种基于图卷积神经网络的古汉语分词框架，通过结合预训练语言模型和图卷积神经网络，将外部知识融合到神经网络模型中来提高分词性能并缓解OOV问题。在《左传》《战国策》和《儒林外史》 3个古汉语分词数据集上的研究结果显示，本文模型提高了3个数据集的分词表现。进一步的研究分析证明，本文模型能够有效地融合词典和N-gram信息；特别是N-gram有助于缓解OOV问题。},
	keywords_zh = {BERT (bidirectional encoder representations from transformers), 古汉语, 图卷积神经网络, 汉语分词, 预训练语言模型},
	language = {zh-CN}
}

@article{wang_evol_2024,
	title = {Evol project: a comprehensive online platform for quantitative analysis of ancient literature},
	volume = {11},
	shorttitle = {Evol project},
	url = {https://www.nature.com/articles/s41599-024-02763-6},
	number = {1},
	urldate = {2025-09-27},
	journal = {Humanities and Social Sciences Communications},
	author = {Wang, Jun and Duan, Siyu and Fu, Binghao and Gao, Liangcai and Su, Qi},
	year = {2024},
	note = {Publisher: Palgrave},
	pages = {1--13},
}

@misc{tang_incorporating_2023,
	title = {Incorporating {Deep} {Syntactic} and {Semantic} {Knowledge} for {Chinese} {Sequence} {Labeling} with {GCN}},
	url = {http://arxiv.org/abs/2306.02078},
	doi = {10.48550/arXiv.2306.02078},
	urldate = {2025-09-27},
	publisher = {arXiv},
	author = {Tang, Xuemei and Wang, Jun and Su, Qi},
	month = jun,
	year = {2023},
	note = {arXiv:2306.02078 [cs]},
}

@article{zh-mo_multimodal_2024,
	title = {Promoting the Development of Ancient Chinese Character Research with Multimodal Large Models},
	title_zh = {以多模态大模型推动中国古文字研究发展},
	volume = {11},
	url = {https://kns.cnki.net/KCMS/detail/detail.aspx?dbcode=CCJD&dbname=CCJDLAST2&filename=YYZV202402004},
	abstract_zh = {人工智能技术与古文字研究的结合已经产生了一批有影响的成果，但几乎都是基于单模态模型。单一模态的智能模型在面对复杂的古文字问题时还存在很大的限制，只能在某些方面起到“辅助”作用，难以独立而完整地解决各种实际问题。文字的形、音、义与各种模态存在不同的关联，汉字的表意属性决定了图像模态和文本模态具有同样重要的作用，中国古文字研究与人工智能的结合须走多模态之路。近年来多模态大模型正在飞速发展，以此为契机来综合性地解决古文字检测、识别、复原等问题，有望取得更好的效果，而文字考释等一些过去难以触及的关键问题也能探索出新的解决路径。},
	language = {zh-CN},
	number = {2},
	urldate = {2025-11-16},
	journal = {Chinese Language Strategy},
	journal_zh = {中国语言战略},
	author = {Mo, Bofeng and Zhang, Chongsheng},
	author_zh = {{莫伯峰} and {张重生}},
	year = {2024},
	keywords_zh = {人工智能, 古文字, 多模态, 甲骨文},
	pages = {37--47},
}

@book{chen_yinque_collected_works_2001,
    title = {Chen Yinque Wenji},
    title_zh = {陈寅恪文集},
    author = {Chen, Yinke},
    author_zh = {陈寅恪},
    publisher = {Sanlian Bookstore},
    publisher_zh = {生活·读书·新知三联书店},
    year = {2001},
    address = {Beijing},
    address_zh = {北京},
	isbn = {9787108033390},
    language = {zh-CN},
}

@article{zh-mo_rejoining_2021,
	title = {Rejoining Oracle Bones through Human-Machine Coupling},
	title_zh = {{AI缀合中的人机耦合}},
	issn = {2096-7365},
	url = {https://kns.cnki.net/KCMS/detail/detail.aspx?dbcode=CJFQ&dbname=CJFDLAST2021&filename=CUWX202101002},
	abstract_zh = {本文结合我们近期的计算机缀合实践,以五组新缀甲骨为例,探讨了AI时代如何利用计算机的优势,并结合专家知识,共同推进古文字研究发展的一些经验。认为AI和专家各有所长,二者需要密切合作,人机耦合是现阶段利用AI技术最为有效的一种方式。},
	language = {zh-CN},
	number = {1},
	urldate = {2025-11-16},
	journal = {Excavated Documents},
	journal_zh = {出土文献},
	author = {Mo, Bofeng and Zhang, Chongsheng and Men, Yi},
	author_zh = {{莫伯峰} and {张重生} and {门艺}},
	year = {2021},
	keywords_zh = {人工智能, 甲骨, 缀合},
	pages = {19--26, 154},
}

@article{zh-mo_computer_2021-1,
	title = {A Discussion about“RI You Ji”in Oracle Bone Inscription Base on a Computer Assistance Rejoining},
	title_zh = {计算机辅助甲骨缀合研讨一则——谈“日有即”},
	url = {https://kns.cnki.net/KCMS/detail/detail.aspx?dbcode=CCJD&dbname=CCJDLAST2&filename=MSDJ202102014},
	abstract_zh = {计算机可以对甲骨缀合起到重要的辅助作用,本文讨论了依靠计算机辅助新缀的一则甲骨:《北珍》435 (《合集》26026)+《北珍》438(《合集》24115),认为缀合复原了一条十分重要的卜辞。新缀卜辞中的“日有即”,与卜辞中的“日月有食”“月有食”“日有戠”以及传世文献中的“日既”等关系密切,是日食的一种类型——日偏食。商人在面对日食时,以人类进食来隐喻这一天文现象。这则新缀合对商代天文和历法研究,都有着非常重要的作用。},
	language = {zh-CN},
	number = {2},
	urldate = {2025-11-16},
	journal = {Folklore, Classics and Chinese Characters},
	journal_zh = {民俗典籍文字研究},
	author = {Mo, Bofeng and Zhang, Zhan},
	author_zh = {{莫伯峰} and {张展}},
	year = {2021},
	keywords_zh = {日食, 甲骨, 缀合, 计算机},
	pages = {169--174, 262--263},
}

@article{zh-wang_computer_2011,
	title = {Research on computer matching of inscriptions on tortoise fragments},
	title_zh = {龟甲类甲骨文碎片计算机辅助缀合研究},
	volume = {32},
	issn = {1000-7024},
	url = {https://link.cnki.net/doi/10.16208/j.issn1000-7024.2011.10.062},
	doi = {10.16208/j.issn1000-7024.2011.10.062},
	abstract_zh = {利用计算机辅助龟甲类甲骨文碎片缀合,建立了碎片数字化处理的基础理论和有关甲骨片的图像预处理、图像轮廓特征以及区域特征提取技术。给出了基于位置数、碎片边界信息、碎片上文字笔画信息、碎片边界上文字信息等5个缀合规则。研究了龟甲类甲骨文碎片计算机辅助缀合系统,仿真结果表明,系统对于待缀合的碎片能自动生成基于"骨版+碎片+特征"三要素的动态疑似目标碎片数据库,在此基础上,通过人机交互根据卜辞类别、卜辞内容、贞人、时期、出土地点等非图片信息进行判断,可以快速辅助用户实现终级缀合。},
	language = {zh-CN},
	number = {10},
	urldate = {2025-11-16},
	journal = {Computer Engineering and Design},
	journal_zh = {计算机工程与设计},
	author = {Wang, Ai-min and Ge, Wen-ying and Zhao, Zhe and Ge, Yan-qiang and Liu, Guo-ying and Li, Qing-sheng},
	author_zh = {{王爱民} and {葛文英} and {赵哲} and {葛彦强} and {刘国英} and {栗青生}},
	year = {2011},
	keywords_zh = {甲骨文, 疑似目标碎片, 缀合, 边界匹配},
	pages = {3570--3573},
}

@article{zh-wang_intelligent_2010,
	title = {Research on Key Technologies of the Computer Aided Rejoining of the Bones/Tortoise Shells with Inscriptions},
	title_zh = {计算机辅助甲骨文缀合关键技术研究},
	volume = {18},
	issn = {1671-4598},
	url = {https://kns.cnki.net/KCMS/detail/detail.aspx?dbcode=CJFQ&dbname=CJFD2010&filename=WHGY201020042},
	abstract_zh = {研究利用计算机辅助甲骨文缀合的有关理论与技术,对于待缀合的碎片自动生成基于"骨版+碎片+特征"三要素的动态疑似目标碎片数据库,在此基础上,基于"时代+字迹+卜辞"三要素研究终级缀合的有关理论和具体实现算法。首先,根据甲骨片图像纹理细节丰富、目标与背景之间差别明显等特点,研究面向甲骨片图像的图像预处理方法;然后,根据甲骨片缀合的实际需求,研究适合于甲骨片缀合的图像轮廓特征、区域特征以及趋势特征提取算法;设计快速高效的甲骨文碎片智能缀合原型系统。},
	language = {zh-CN},
	number = {20},
	urldate = {2025-11-16},
	journal = {Journal of Wuhan University of Technology},
	journal_zh = {武汉理工大学学报},
	author = {Wang, Ai-min and Zhong, Luo and Ge, Yan-qiang and Liu, Guo-ying},
	author_zh = {{王爱民} and {钟珞} and {葛彦强} and {刘国英}},
	year = {2010},
	keywords_zh = {甲骨文, 算法, 缀合, 计算机, 边界匹配},
	pages = {194--199},
}

@article{zh-wang_computer_2010-2,
	title = {System designation for computer aided rejoining of bones/tortoise shells with inscriptions based on contour matching},
	title_zh = {甲骨文计算机辅助缀合系统设计},
	volume = {46},
	issn = {1002-8331},
	url = {https://kns.cnki.net/KCMS/detail/detail.aspx?dbcode=CJFQ&dbname=CJFD2010&filename=JSGG201021019},
	abstract_zh = {计算机辅助甲骨文碎片缀合,是整理甲骨的一种先进技术。研究了甲骨片图像的轮廓信息的提取与轮廓跟踪算法、轮廓片段特征向量提取算法,建立了甲骨文碎片数据库,研制了基于边界匹配的甲骨文缀合辅助系统,选定待缀合的甲骨碎片后,该系统可以自动生成疑是目标甲骨碎片的动态数据库,甲骨文专家只需要基于"备选甲骨碎片数据库"通过人机交互来实现甲骨文缀合。},
	language = {zh-CN},
	number = {21},
	urldate = {2025-11-16},
	journal = {Computer Engineering and Applications},
	journal_zh = {计算机工程与应用},
	author = {Wang, Ai-min and Liu, Guo-ying and Ge, Wen-ying and Zhou, Hongyu and Wang, Dinglei},
	author_zh = {{王爱民} and {刘国英} and {葛文英} and {周宏宇} and {王丁磊}},
	year = {2010},
	keywords_zh = {甲骨文, 缀合, 计算机, 边界匹配},
	pages = {59--62},
}

@article{zh-li_computer_1996,
	title = {A Newly Discovery on words processing Technology —The Design of Pictographic Code to Inscriptions on Bones or Tortoise Shells},
	title_zh = {计算机文字信息处理技术新探——甲骨文象形码设计方案},
	issn = {1003-0077},
	url = {https://kns.cnki.net/KCMS/detail/detail.aspx?dbcode=CJFQ&dbname=CJFD9697&filename=MESS199603002},
	abstract_zh = {作为计算机中文信息处理技术的有机组成部队的输入法,常常被称作“瓶颈”。甲骨文输入法是这个“瓶颈”的最困难部分。本方案在此建立和阐述了一种甲骨文输入法——象形码输入法。在考虑了文字信息处理中关于部件系统的规律后,文章对甲骨文形体架构特性作了分析;利用现代编码规律,给出了甲骨文部件(字根)形成和码元确定的规则。全文有完整的码元、字根表、例示和详细规则说明,同时也显示了象形码的文字学意义与操作简便性共存的特点。},
	language = {zh-CN},
	number = {3},
	urldate = {2025-11-16},
	journal = {Journal of Chinese Information Processing},
	journal_zh = {中文信息学报},
	author = {Li, Jimin},
	author_zh = {{李继明}},
	year = {1996},
	pages = {18--29},
}

@article{zh-mo_initial_2021,
	title = {A Preliminary Test of Artificial Intelligence Simulating the Induction of Citation Examples},
	title_zh = {人工智能模拟辞例归纳的初步测试},
	volume = {12},
	issn = {1674-8506},
	url = {https://kns.cnki.net/KCMS/detail/detail.aspx?dbcode=CJFQ&dbname=CJFDLAST2021&filename=HYWX202103021},
	abstract_zh = {古文字考释中的辞例归纳法,其实是综合了经验和理性两个方面共同作用的一种词义推定方法。人工智能语言模型现在主要模拟了人类经验主义的方法,并在日常语言处理方面取得了比较好的效果。如果将此类模型运用于古文字领域来模拟辞例归纳,也定会有所助益。我们基于Bert模型,用《四库全书》作为特定语料对模型进行了训练。以《上博简》(1-9)中2103个字为测试对象,模拟专家的部分辞例归纳能力,预测被遮蔽起来的文字。在总数23157的备选字符中,前300预测正确率达到59\%,前100预测正确率达到46\%,前50预测正确率达到38\%,前10预测正确率达到25\%,前5预测正确率达到20\%。可见,人工智能在古文字领域也具有类似人脑凭借语言经验进行辞例归纳的能力。同时,结果也提示,必须结合理性主义方法,才能实现完整的辞例归纳能力,建立相关的知识库必不可少。},
	language = {zh-CN},
	number = {3},
	urldate = {2025-11-16},
	journal = {Studies of Chinese Language and Literature},
	journal_zh = {汉语言文学研究},
	author = {Mo, Bofeng and Qiu, Weiqi and Xie, Zecheng},
	author_zh = {{莫伯峰} and {邱炜琦} and {谢泽澄}},
	year = {2021},
	keywords_zh = {人工智能, 古文字, 辞例归纳},
	pages = {128--135},
}

@article{zh-mo_daily_2025,
	title = {The Daily Scientific Research of Oracle Bone Scholars},
	title_zh = {甲骨学家的科研日常},
	issn = {0529-150X},
	url = {https://kns.cnki.net/KCMS/detail/detail.aspx?dbcode=CJFN&dbname=CJFDLASN2025&filename=ZSLL202507013},
	abstract_zh = {{\textless}正{\textgreater}自1899年殷墟甲骨被发现以来，甲骨学家的主要工作就是与这些3000多年前的龟甲和牛骨打交道——搜寻流散于世界各地的甲骨材料，破解甲骨上那些神秘的文字符号，还原隐藏于文字背后的殷商历史原貌。甲骨学家的科研工作可谓是一场寻宝探秘之旅。以王国维、孙诒让这些清末民初国学大师为代表的初代甲骨学家，确实曾经一手握着毛笔，一手持着甲骨拓本来进行研究。然而如今，网络化、数字化、智能化正逐步改变着甲骨学家的科研日常。},
	language = {zh-CN},
	number = {7},
	urldate = {2025-11-16},
	journal = {Knowledge is Power},
	journal_zh = {知识就是力量},
	author = {Mo, Bofeng},
	author_zh = {{莫伯峰}},
	year = {2025},
	pages = {36--37},
}

@inproceedings{hua_bird_2025,
  title     = {BIRD: Bronze Inscription Restoration and Dating},
  author    = {Hua, Wenjie and Nguyen, Hoang H. and Ge, Gangyan},
  booktitle = {Proceedings of the 2025 Conference on Empirical Methods in Natural Language Processing},
  year      = {2025},
  publisher = {Association for Computational Linguistics}
}

\end{document}